\documentclass[10pt,twocolumn,letterpaper]{article}

\usepackage{cvpr}
\usepackage{times}
\usepackage{epsfig}
\usepackage{graphicx}
\usepackage{amsmath}
\usepackage{amssymb}
\usepackage{color}
\usepackage{subfigure}
\usepackage{stfloats}

\usepackage[pagebackref=true,breaklinks=true,letterpaper=true,colorlinks,bookmarks=false,citecolor=blue,linkcolor=blue]{hyperref}

\cvprfinalcopy 

\def\cvprPaperID{1173} 

\ifcvprfinal\fi
\begin{document}

\title{Bridging Stereo Matching and Optical Flow via Spatiotemporal Correspondence}

\author{Hsueh-Ying Lai$^1$ \quad Yi-Hsuan Tsai$^2$ \quad Wei-Chen Chiu$^1$\\
$^1$National Chiao Tung University, Taiwan \qquad $^2$NEC Laboratories America\\
}

\maketitle

\begin{abstract}
Stereo matching and flow estimation are two essential tasks for scene understanding, spatially in 3D and temporally in motion.
Existing approaches have been focused on the unsupervised setting due to the limited resource to obtain the large-scale ground truth data.
To construct a self-learnable objective, co-related tasks are often linked together to form a joint framework.
However, the prior work usually utilizes independent networks for each task, thus not allowing to learn shared feature representations across models. 
In this paper, we propose a single and principled network to jointly learn spatiotemporal correspondence for stereo matching and flow estimation, with a newly designed geometric connection as the unsupervised signal for temporally adjacent stereo pairs.
We show that our method performs favorably against several state-of-the-art baselines for both unsupervised depth and flow estimation on the KITTI benchmark dataset.

\end{abstract}

\section{Introduction}\label{sec:intro}
Reconstructing 3D motion from the real-world visual data has long been a fundamental problem in computer vision and is substantial for numerous applications such as robotics, virtual/augmented reality, and autonomous driving.
Among the tasks of understanding 3D motion, two of the most commonly studied scenarios are optical flow estimation and stereo matching for depth estimation.
Generally, the motion in 3D after projection into the image plane of a camera stands for the optical flow between two consecutive frames in a video, while the 3D structure captured by two horizontally displaced cameras builds the stereo rig as the binocular vision system of human eyes.
Thus, the estimation of optical flow and stereo matching, which discover the pixel displacement across temporally adjacent frames and stereo pairs, provide crucial access to the 3D information.

Recently, deep learning-based approaches have shown tremendous improvement for both optical flow estimation and stereo matching in the supervised learning setting {\cite{eigen2014depth, Ladicky2014PullingTO, Liu2016LearningDF, Fischer2015FlowNetLO, Ilg2017FlowNet2E, Gadot2016PatchBatchAB, Gney2016DeepDF, Ranjan2017OpticalFE}}.
However, these methods usually rely on large-scale datasets with ground truths, but such annotating efforts are significantly expensive, especially in forms of pixel-wise displacement for optical flow and stereo matching.
For eliminating
limitation of datasets and potential issues such as poor model generalization across various scenes, several approaches are proposed recently to explore the unsupervised learning frameworks~\cite{zhou2017unsupervised, Meister:2018:UUL, monodepth17}.
\begin{figure}[t]
\centering
  \includegraphics[width=0.48\textwidth]{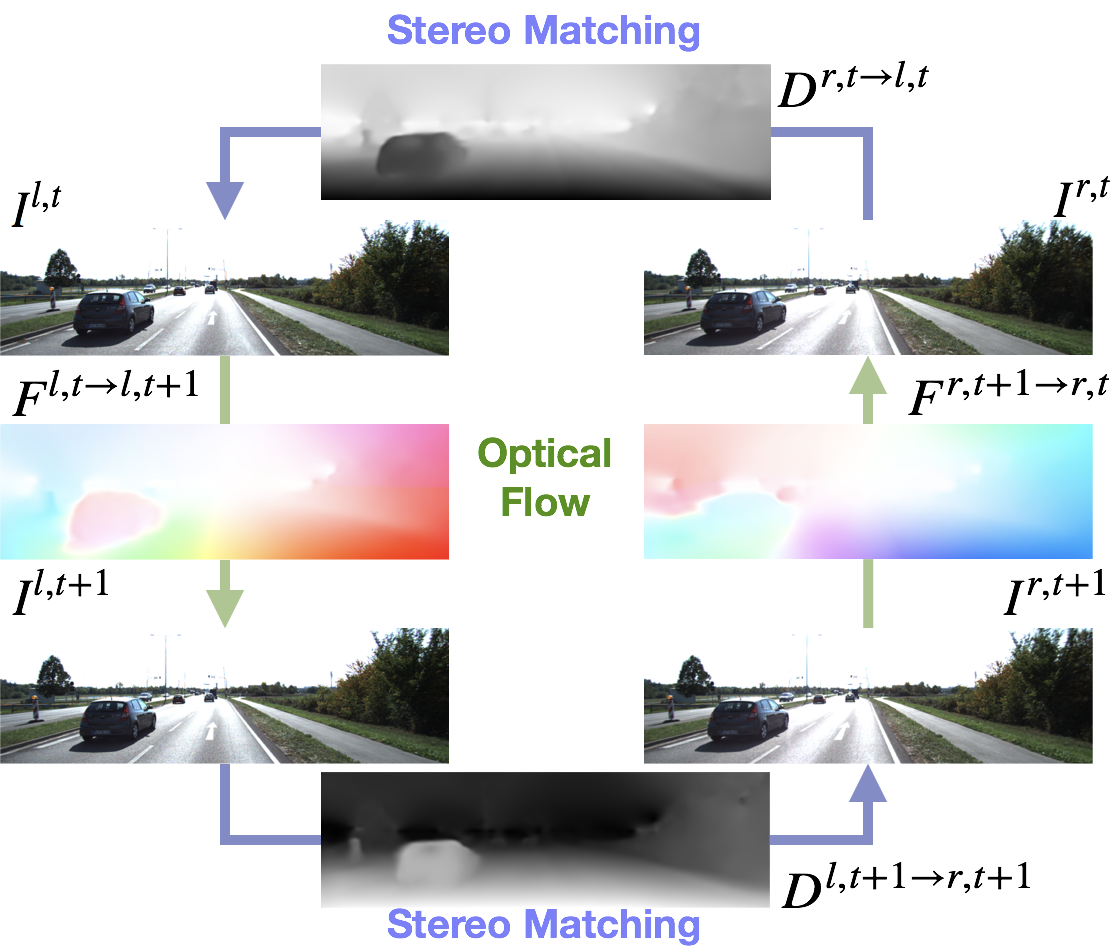}
  \caption{Using temporally adjacent stereo pairs as input, our model can estimate the correspondence maps of each pair via geometric connections, thus bridging stereo matching and optical flow through multiple reconstruction, forming a cycle.}
  \label{fig:teaser}
\end{figure}

In the unsupervised learning setting, a common practice is to relate different tasks (e.g., optical flow, depth estimation, or camera pose estimation) and utilize photometric consistency to measure the pixel correspondences across frames {\cite{2018arXiv180509806R, 2018arXiv180610556Y, yin2018geonet, zhou2017unsupervised, zou2018dfnet}}.
Nevertheless, existing approaches utilize separate networks for each task, and thus the feature representations are not effectively shared across tasks.
In this paper, we argue that there should exist a principled model, which is capable of learning joint representations for tasks that are highly co-related.
Although the properties of pixel correspondence used in stereo matching and optical flow estimation are slightly different, as the former considers the horizontal offset while the later has movement in both horizontal and vertical directions, the common goal is obviously shared (i.e., finding pixel correspondences).
By taking advantages of such a correlation, we propose to design a single network for simultaneously estimating optical flow and stereo matching, and show that these two tasks are beneficial to each other via learning shared feature representations.
Moreover, we construct an unsupervised learning framework with modelling the geometric connections between both tasks based on temporally adjacent stereo pairs (as shown in Figure~\ref{fig:teaser}), in which this type of data is easily accessible as the popularity of stereo video cameras.
We design a warping function that considers the consistency across adjacent video frames, and sequentially feed the training data both from flow and stereo pairs to meet the designed geometric constraints.
Extensive experiments are conducted on both KITTI2012~\cite{Geiger2012CVPR} and KITTI2015~\cite{Menze2015ISA} benchmark dataset to evaluate the effectiveness of the proposed method and show favorable performance against several state-of-the-art algorithms.
In addition, we sequentially demonstrate the mutual
benefit of jointly learning both tasks of optical flow estimation and stereo matching, successfully showing the improvement via utilizing the proposed geometric connections built upon stereo video data.
The main contributions of the paper are summarized as the follows:
\begin{itemize}
    \item We propose a single and principled network for joint estimating optical flow and stereo matching to account for their shared representations, in which the common goal is to find pixel correspondence across images.

    \item We introduce geometric constraints during the joint learning process, which provides an effective signal for modeling the consistency (i.e., spatiotemporal correspondence) across two tasks and is then utilized as an objective for unsupervised training.

    \item We develop an efficient training scheme for the joint optimization on two tasks within a single framework and show that both tasks benefit each other.
\end{itemize}

\section{Related Works}\label{sec:related}
We organize and discuss related approaches, including stereo matching, depth estimation, optical flow estimation, and the joint framework of them.

\paragraph{Unsupervised Learning of Depth Estimation.} Stereo matching for depth estimation has been a classical computer vision problem for decades. Prior to the recent renaissance of deep learning, many approaches are proposed to tackle this problem based on diverse strategies, such as hand-crafting feature descriptors for matching local regions across frames, or formulating stereo matching upon a graphical model and resolving it by complicated energy minimization. With large annotated datasets are available (e.g., KITTI~\cite{Geiger2012CVPR}) in recent years, better matching functions to measure the similarity between image patches are learnt by deep neural networks \cite{luo2016efficient, zbontar2016stereo, chang2018pyramid} which obtain significant boost in performance. Simultaneously, estimating depth directly from monocular images based on deep models in the supervised learning manner is also widely explored~\cite{eigen2014depth, Liu2016LearningDF}. However, the requirement for training data with ground truths is expensive to meet, and thus the unsupervised learning scheme~\cite{zhou2017unsupervised, monodepth17, 2017arXiv170407813Z, 2018arXiv180302612L} is popularly adopted. Here we review several of them as follows.

Godard \etal \cite{monodepth17} learn to estimate disparity maps which are used to warp between images in a stereo pair for optimizing objectives of left-right consistency. Instead of exploring the pixel correspondence within stereo pairs, given a video sequence, \cite{2017arXiv170407813Z} jointly estimates both the monocular depth of each frame as well as the camera motion such that consecutive frames can be reconstructed between each others, and are used for evaluating photometric consistency as loss functions.
In \cite{2018arXiv180302612L}, the authors combine the concept of monocular depth estimation and stereo matching, where binocular views in a stereo pair are first synthesized by using the depth map estimated from the monocular image. Then the stereo matching network is applied to produce the final depth estimation. Typically, these methods attempt to regress depth map solely from monocular inevitably depends on the quality of training data and hardly generalize to unseen scenes. In contrast, the models for stereo matching concentrate on learning to match pixels between images and thus have better generalizability, in which we aim to address the same stereo matching task in this paper. In the work of Zhou \etal \cite{zhou2017unsupervised}, the authors propose to learn stereo matching via iterative left-right consistency check. Godard \etal \cite{monodepth17} also extend their monocular depth estimation framework to perform stereo matching and obtain better performance with respect to its monocular version.

\paragraph{Unsupervised Learning of Optical Flow.} The research works addressing optical flow estimation follow the same evolution as the ones for depth estimation, starting from conventional methods~\cite{fleet2006optical,fortun2015optical}, advancing to deep learning models based on supervised setting~\cite{Fischer2015FlowNetLO, Ilg2017FlowNet2E}, and then exploring unsupervised learning approaches~\cite{jjyu2016unsupflow, AAAI1714388, Meister:2018:UUL}. When unsupervised learning of optical flow are first introduced in FlowNet-Simple \cite{jjyu2016unsupflow} and DSTFlow \cite{AAAI1714388}, they utilize the similar objectives of photometric consistency across frames and local smoothness in the estimated flow map. However, these works do not take the severe occlusion issue into consideration when there are objects with large movement. In order to resolve the artifacts resulted from the warping operation, \cite{2017arXiv171105890W, Meister:2018:UUL, Janai2018ECCV} handle regions of occlusion by analyzing the inconsistency between forward and backward flow maps. \cite{Meister:2018:UUL} further replaces the typical L1 loss with the ternary census transform for measuring photometric consistency, providing more reliable constancy assumption in realistic situations. Moreover, \cite{Janai2018ECCV} 
advance the optical flow estimation and occlusion handling by explicitly reasoning over multiple consecutive frames within a time window.

\paragraph{Joint Learning Framework of Depth and Optical Flow.} Recently, numerous works have been proposed to jointly learn both depth and optical flow estimation models via employing geometric relations between flow, depth, and camera poses. In~\cite{2017arXiv170407813Z}, based on the assumption of rigid scenes, the pixel correspondence between temporally adjacent frames caused by camera movement is derived from the estimates on both monocular depth and camera poses, and thus it becomes the key to define objectives for joint training.
GeoNet~\cite{yin2018geonet} follows the similar idea as~\cite{2017arXiv170407813Z} but particularly introduces non-rigid motion localizer to handle moving objects in the optical flow map. Yang \etal \cite{2018arXiv180610556Y} explicitly disentangle the dynamic objects from static background in a video based on a motion network, and carefully model it together with depth, flow, and camera pose estimation by using geometric constraints. The occlusion mask as well as 3D motion maps for dynamic and static regions can thus be obtained. DF-net \cite{zou2018dfnet} especially leverages the geometric consistency between the estimated flow from optical flow model and the synthetic 2D optical flow obtained from estimates of depth and camera motion, in which it shows benefits for simultaneously
training monocular depth prediction and optical flow estimation networks.
Along the same track of unsupervised learning but unlike the aforementioned research works where separate networks are learned for each task, our proposed method tackles stereo matching and optical flow estimation within a single and principled network, and relates them through geometric connections built upon temporally adjacent stereo pairs.

\section{Proposed Method}\label{sec:method}
In this section, we first describe the overall structure of how we construct the geometric relations among stereo videos.
Second, we introduce each component of the proposed method, including unsupervised loss functions shared across stereo matching and flow estimation, a newly proposed 2-Warp loss that measures the consistency between two tasks, and occlusion handling for flow estimation.
\begin{figure}[t]
\centering
  \includegraphics[width=0.4\textwidth]{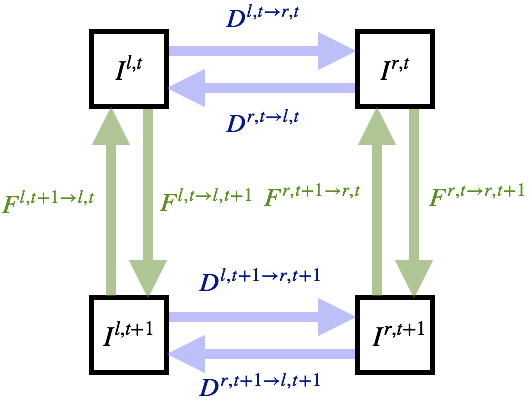}
  \caption{The relation of bridging stereo pairs and consequent frames. We can estimate the correspondence maps of any directions based on the input pairs and their reconstruction direction.}
  \label{fig:corr}
\end{figure}
\begin{figure*}[t]
\centering
  \includegraphics[width=0.7\linewidth]{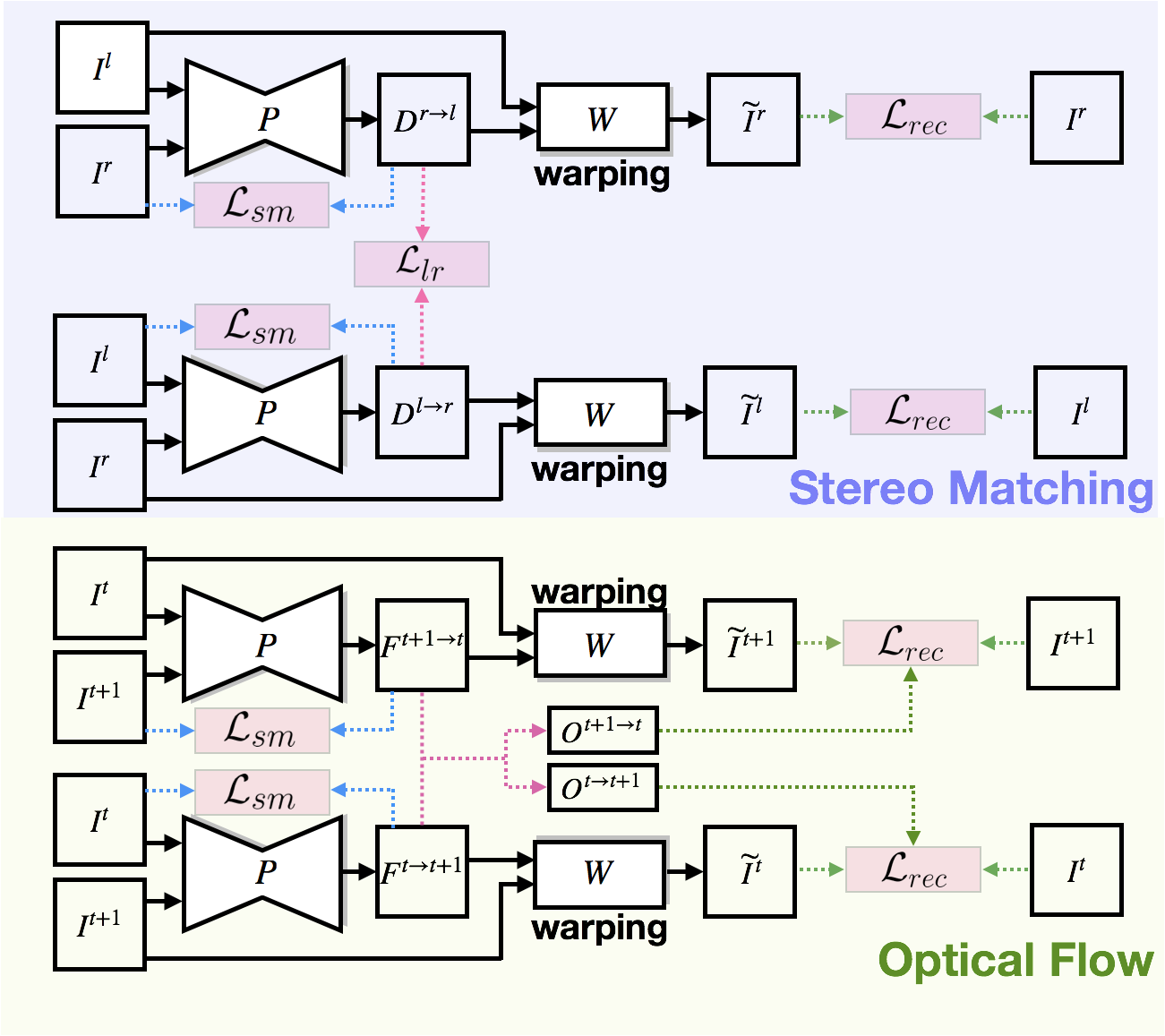}
  \caption{Overall structure of our method. Our framework consists of a single model $P$ that estimates dense correspondence maps based on the order of two input images for both stereo matching and optical flow.
  Each pair can be fed into $P$ but in a different image order (e.g., ($I_l, I_r$) and ($I_r, I_l$)), and thus two reconstruction loss $\mathcal{L}_{rec}$ are able to be optimized based on two warping functions $W$ obtained from each pair.
  Between these two tasks, two difference are: (1) we apply left-right consistency $\mathcal{L}_{lr}$ to stabilize the stereo matching part only; (2) occlusion map derived from the correspondence maps of two opposite directions is adopted on the reconstruction loss for solving the largely occluded area for optical flow only.}
  \label{fig:model}
  \vspace{-1mm}
\end{figure*}
\subsection{Overall Structure}
\label{sec:overall}

As motivated previously that both optical flow estimation and stereo matching aim to find pixel correspondences across images, our goal is to learn a single and principled network for these two tasks in an unsupervised learning manner with exploiting their geometric relations stemmed from stereo videos. Figure~\ref{fig:model} illustrates the framework of the proposed method, which will be detailed in the following subsections.

The network $P$ in our method is based on the model used in \textit{Monodepth}~\cite{monodepth17}, which is now extended from its original usage of monocular depth estimation to take two input frames and output both horizontal and vertical offsets for pixel correspondences across input frames. Assuming two temporally adjacent stereo pairs are given as $\left\{ I^{l,t}, I^{r,t}, I^{l,t+1}, I^{r,t+1}\right\}$ where the superscripts $l, r$ denote left and right frames in a stereo pair respectively, and $t, t+1$ indicate their temporal indexes. Our network $P$ is able to perform stereo matching to obtain the forward pixel correspondence $D^{l,t \rightarrow r,t}$ from $I^{l,t}$ to $I^{r,t}$ as well as the backward one $D^{r,t \rightarrow l,t}$ from $I^{r,t}$ to $I^{l,t}$ :
\begin{equation}
\begin{aligned}
D^{l,t \rightarrow r,t} &= P(I^{l,t}, I^{r,t})\\
D^{r,t \rightarrow l,t} &= P(I^{r,t}, I^{l,t})\\
\end{aligned}
\end{equation}
Likewise, for another stereo pair at time $t+1$, we obtain:
\begin{equation}
\begin{aligned}
D^{l,t+1 \rightarrow r,t+1} &= P(I^{l,t+1}, I^{r,t+1})\\
D^{r,t+1 \rightarrow l,t+1} &= P(I^{r,t+1}, I^{l,t+1})\\
\end{aligned}
\end{equation}
The forward/backward optical flow maps on the left and right views can also be estimated using our network:
\begin{equation}
\begin{aligned}
F^{l,t \rightarrow l,t+1} &= P(I^{l,t}, I^{l,t+1})\\
F^{l,t+1 \rightarrow l,t} &= P(I^{l,t+1}, I^{l,t})\\
F^{r,t \rightarrow r,t+1} &= P(I^{r,t}, I^{r,t+1})\\
F^{r,t+1 \rightarrow r,t} &= P(I^{r,t+1}, I^{r,t})\\
\end{aligned}
\end{equation}
The overall relations are shown in Figure \ref{fig:corr}. With these pixel correspondences, we aim to reconstruct a frame given its counterpart of a stereo pair or its temporal adjacency, based on a warping function $W$. For instance, frame $I^{r,t}$ can be reconstructed as:
\begin{equation}
\tilde{I}^{r,t} = W(I^{l,t}, D^{r,t \rightarrow l,t}),
\end{equation}
from its corresponding left view $I^{l,t}$ and the backward stereo matches $D^{r,t \rightarrow l,t}$. Similarly, $I^{l,t}$ can be reconstructed as:
\begin{equation}
\tilde{I}^{l,t} = W(I^{l,t+1}, F^{l,t \rightarrow l,t+1}),
\end{equation} from its next frame $I^{l,t+1}$ via the flow $F^{l,t \rightarrow l,t+1}$.
For simplicity, we skip listing here for other combinations across frames, which should be easily derivable.
\subsection{Occlusion Estimation for Optical Flow}
\label{sec:occ}
Before introducing the designed unsupervised loss functions in our framework, we describe first how we tackle the common occlusion issue for flow estimation.
During training, there would be some occluded regions only visible at frame $t$ but having no corresponding pixels at frame $t+1$, as the camera or objects may have large movement. This causes the inconsistent warping process in appearance between the reconstructed image and the target one.

In order to deal with the occlusion issue, we utilize the forward-backward consistency check~\cite{sundaram2010dense,2017arXiv171105890W, zou2018dfnet} to localize the potentially occluded regions. 
More precisely, applying warping operation on a backward map by its corresponding forward map, e.g., $W(F^{l,t+1 \rightarrow l,t}, F^{l,t \rightarrow l,t+1}$), ideally could reconstruct the forward map with a negative sign in non-occluded regions. To this end, we follow the technique as used in \cite{Meister:2018:UUL} and by taking the pair of $\left \{I^{l,t}, I^{l,t+1} \right \}$ as an example, pixels are considered as occluded while the criterion below is violated:
\small
\begin{equation}
\begin{aligned}
|&F^{l,t \rightarrow l,t+1} + W(F^{l,t+1 \rightarrow l,t}, F^{l,t \rightarrow l,t+1})|^2 \\
&< {\alpha}_1(|F^{l,t \rightarrow l,t+1}|^2 + |W(F^{l,t+1 \rightarrow l,t}, F^{l,t \rightarrow l,t+1})|^2) + {\alpha}_2,
\end{aligned}
\end{equation}
\normalsize
where the hyper-parameters $\alpha_1$ and $\alpha_2$ are set to $0.01$ and $0.5$ respectively. An occlusion map $O$ is then obtained by setting 0 to those occluded regions and 1, otherwise.

\begin{figure*}[!t]
    \subfigure[]{
    \label{Fig.sub.1}
    \includegraphics[height=0.24\textheight]{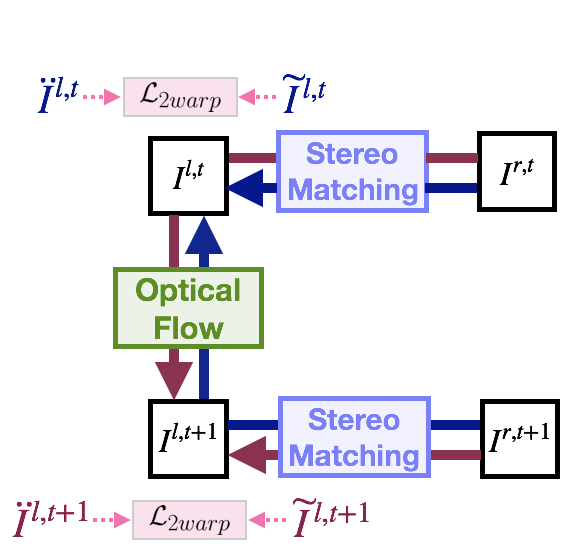}}
    \subfigure[]{
    \label{Fig.sub.2}
    \includegraphics[height=0.24\textheight]{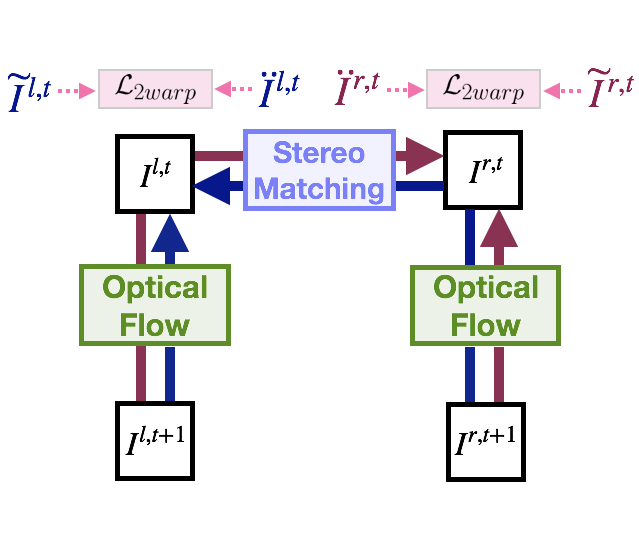}}
    \subfigure[]{
    \label{Fig.sub.3}
    \includegraphics[height=0.24\textheight]{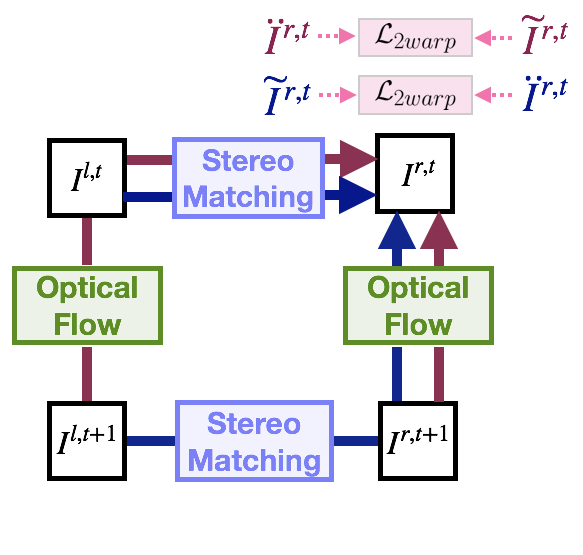}}
    \caption{Our proposed 2-warp modules. The arrows indicate the warping direction and a 2-warp reconstruction loss is performed when the arrows with the same color meet, forcing reconstructed images via the 2-warp operations to be consistent. Here, we introduce three types of 2-warp functions and will discuss them in the section of experiments.
    }
    \label{fig:2warp}
\end{figure*}
\subsection{Unsupervised Loss Functions}
One key factor to make the proposed unsupervised method work is to design plausible loss functions that can exploit various connection across video frames.
In the following, we sequentially introduce the utilized loss functions, including self-supervised reconstruction loss, smoothness loss, left-right consistency loss, and 2-Warp consistency loss to model the relation across stereo videos.
Here, we use the pair of $\left \{I^{l,t}, I^{l,t+1} \right \}$ as an example for explanation and all the loss functions apply to both stereo/flow pairs, unless stated specifically.
The overall structure of the proposed framework and loss functions is presented in Figure~\ref{fig:model}.

\paragraph{Reconstruction Loss.}
The reconstruction loss $\mathcal{L}_{rec}$ is similar to the one used in Monodepth~\cite{monodepth17} but with occlusion-aware constraints. The loss is the weighted sum of SSIM-based loss and L1 loss which compares ${I}^{l,t}$ and its reconstruction $\tilde{I}^{l,t}$:
\begin{equation} \label{eq:photo}
\begin{aligned}
\mathcal{L}^{l,t \rightarrow l,t+1}_{rec}
= \frac{1}{\sum_{i,j}O}[\sum_{i,j}(\alpha\frac{1 - SSIM({I}^{l,t}_{ij}, \tilde{I}^{l,t}_{ij})}{2}\\
+ (1 - \alpha)|{I}^{l,t}_{ij} - \tilde{I}^{l,t}_{ij}|) \cdot O],
\end{aligned}
\end{equation}
where $O$ is the occlusion map derived from Section~\ref{sec:occ}, subscript $i,j$ indicates pixel coordinates, and $\alpha$ denotes the weights between SSIM and L1 loss. Since our occlusion maps are only used in the image pairs for flow estimation, all the elements in the occlusion map would be equal to $1$ when $\mathcal{L}_{rec}$ is applied on image pairs for stereo matching.
\paragraph{Smoothness Loss.}
For the smoothness loss $\mathcal{L}_{sm}$, we adopt the formulation introduced in \cite{2017arXiv171105890W} which encourages the correspondence maps to be locally smooth but also maintains edges that should be aligned with the structure of images:
\begin{equation}
\mathcal{L}^{l,t \rightarrow l,t+1}_{sm} = \frac{1}{N}\sum_{i,j}\sum_{d\in(x,y)}
|\partial^2_dF^{l,t \rightarrow l,t+1}|e^{-\beta|\partial_d{I}^{l,t}_{ij}|}
\end{equation}
where $\beta$ denotes the edge-weighted hyperparameter.
Here we adopt the second-order and the first-order derivatives on the correspondence map and the image, respectively.
\paragraph{Left-right Correspondence Consistency Loss.}
To improve the accuracy of correspondence map estimation as well as balance the performance of left-right estimation, we not only check the consistency of left-right reconstruction but also check left-right correspondence.
Similar to the occlusion detection, our left-right consistency loss $\mathcal{L}_{lr}$ is derived from reconstructing the correspondence map pair by warping each other and compute the absolute L1 difference loss. Following~\cite{2018arXiv181003654W}, this consistency term is only adopted on stereo pairs:
\small
\begin{equation}
\mathcal{L}^{l,t \rightarrow r,t}_{lr} = \frac{1}{N}\sum_{i,j}|D^{l,t \rightarrow r,t} + W(D^{r,t \rightarrow l,t}, D^{l,t \rightarrow r,t})|
\end{equation}
\normalsize
\paragraph{2-Warp Consistency Loss.}
To reinforce the structure of stereo matching and optical flow estimation, we introduce a new 2-Warp consistency loss. That is, we warp an image twice through both the optical flow and stereo sides. Figure~\ref{fig:2warp} presents three types of the possible 2-warp operations that we investigate. We will introduce the details of the first one as follows, while the others can be derived similarly.

Following previous works of depth estimation, we do not apply occlusion maps on stereo pairs, so that we could easily derive the 2-warp occlusion maps from estimated flow maps.
To reconstruct $I^{r,t}$ from $I^{l,t+1}$ via $I^{l,t}$, the occlusion map and the 2-warp reconstructed image are written as:
\begin{equation}
O^{r,t \rightarrow l,t+1} = W(O^{l,t \rightarrow l,t+1}, D^{r,t+1 \rightarrow l,t+1}).
\end{equation}
\begin{equation}
\ddot{I}^{r,t} = W(W(I^{l,t+1}, F^{l,t \rightarrow l,t+1}), F^{r,t \rightarrow l,t}).
\end{equation}
The occlusion regions between the stereo pairs at time $t$ is the area where objects occlude the background at time $t+1$, so the occlusion area can be mapped by $D^{l,t+1 \rightarrow r,t+1}$. Therefore, warping $O^{l,t \rightarrow l,t+1}$ by $D^{r,t+1 \rightarrow l,t+1}$ as our 2-warp occlusion map is valid. Similar to~\eqref{eq:photo}, we could apply the occlusion-aware reconstruction loss between reconstructed $\ddot{I}^{r,t}$ from $I^{l,t+1}$ via $I^{l,t}$ and the reconstructed $\tilde{I}^{r,t}$ directly from $I^{r,t+1}$, as illustrated in Figure~\ref{fig:2warp}(a).
\begin{equation} \label{eq:2warp}
\begin{aligned}
\mathcal{L}^{r,t \rightarrow l,t+1}_{2warp}
= \frac{1}{\sum_{i,j}O^{r,t \rightarrow l,t+1}}[\sum_{i,j}(\alpha\frac{1 - SSIM(\ddot{I}^{r,t}_{ij}, \tilde{I}^{r,t}_{ij})}{2} \\
+ (1 - \alpha)|\ddot{I}^{r,t}_{ij} - \tilde{I}^{r,t}_{ij}|) \cdot O^{r,t \rightarrow l,t+1}].
\end{aligned}
\end{equation}

\paragraph{Total Loss.} The total loss of the proposed framework is:
\begin{equation}
\mathcal{L}_{total} = \mathcal{L}_{rec} + \lambda_{sm}\mathcal{L}_{sm} + \lambda_{lr}\mathcal{L}_{lr} + \lambda_{2warp}\mathcal{L}_{2warp}
\end{equation}
We note that, all these terms except 2-Warp consistency, have its mirrored counterpart at each scale for the multi-scale estimation as in Monodepth \cite{monodepth17}.

\begin{table*}[!t]
    \centering
    \renewcommand{\arraystretch}{1.2}
    \setlength{\tabcolsep}{2pt}
    \caption{Quantitative evaluation of the depth estimation task on KITTI 2015 stereo set. Our results are capped between 0-80 meters. Our full model includes settings with three types of 2-warp operations from Figure \ref{fig:2warp} and full-1,2,3 correspond to Figure \ref{Fig.sub.1}, \ref{Fig.sub.2} and \ref{Fig.sub.3} respectively. Using stereo pairs during training/testing is also indicated in the table.}
    \vspace{1mm}
    \begin{tabular}{lcc|lccc|ccc}
    \hline 
        Method & Train & Test & \multicolumn{4}{c|}{Lower the better} & \multicolumn{3}{c}{Higher the better} \\
        & Stereo & Stereo & Abs Rel & Sq Rel & RMSE & RMSE log & $\delta < 1.25$ & $\delta < 1.25^2$ & $\delta < 1.25^3$ \\
    \hline
        Wang et al. \cite{Wang_2018_CVPR} &&& 0.148 & 1.187 & 5.496 & 0.226 & 0.812 & 0.938 & 0.975 \\
        Godard et al. \cite{monodepth17} & \checkmark && 0.097 & 0.896 & 5.093 & 0.176 & 0.879 & 0.962 & 0.986 \\
        Yang et al. \cite{2018arXiv180610556Y} & \checkmark && 0.099 & 0.986 & 6.122 & 0.194 & 0.860 & 0.957 & 0.986 \\
    \hline
        Godard et al. \cite{monodepth17} & \checkmark & \checkmark & 0.068 & 0.835 & 4.392 & 0.146 & 0.942 & 0.978 & 0.989 \\
        Ours (stereo only) & \checkmark & \checkmark & 0.078 & 0.811 & 4.700 & 0.174 & 0.918 & 0.965 & 0.983 \\
        Ours (flow + stereo) & \checkmark & \checkmark & 0.0653 & 0.819 & 4.268 & 0.151 & 0.946 & \textbf{0.979} & \textbf{0.990} \\
        Ours (full-1) & \checkmark & \checkmark & 0.0631 & 0.756 & 4.207 & 0.147 & 0.947 & \textbf{0.979} & \textbf{0.990} \\
        Ours (full-2) & \checkmark & \checkmark & \textbf{0.0620} & \textbf{0.747} & \textbf{4.113} & \textbf{0.146} & \textbf{0.948} & \textbf{0.979} & \textbf{0.990} \\
        Ours (full-3) & \checkmark & \checkmark & 0.0630 & 0.773 & 4.195 & 0.147 & 0.947 & \textbf{0.979} & \textbf{0.990} \\
    \hline
    \end{tabular}
    \label{tab:kitti}
\end{table*}
\begin{table*}[!t]
    \centering
    \renewcommand{\arraystretch}{1.2}
    \setlength{\tabcolsep}{2pt}
    \caption{Quantitative evaluation of the depth estimation task on the KITTI raw dataset split by Eigen et al.~\cite{eigen2014depth}. All results are cropped based on the setting in \cite{garg2016unsupervised}. Using stereo pairs during training/testing or supervised data is indicated in the table.}
    \vspace{1mm}
    \begin{tabular}{lccc|cccc|ccc}
    \hline 
        Method & Train & Test & Super- & \multicolumn{4}{c|}{Lower the better} & \multicolumn{3}{c}{Higher the better} \\
        & Stereo & Stereo & vised & Abs Rel & Sq Rel & RMSE & RMSE log & $\delta < 1.25$ & $\delta < 1.25^2$ & $\delta < 1.25^3$ \\
    \hline
        Eigen et al. \cite{eigen2014depth} &&& \checkmark & 0.203 & 1.548 & 6.307 & 0.282 & 0.702 & 0.890 & 0.958 \\
        Godard et al. \cite{monodepth17} & \checkmark &&& 0.114 & 0.898 & 4.935 & 0.206 & 0.861 & 0.949 & 0.976 \\
        Yang et al. \cite{2018arXiv180610556Y} & \checkmark &&& 0.114 & 1.074 & 5.836 & 0.208 & 0.856 & 0.939 & 0.976 \\
        Zhou et al. \cite{2017arXiv170407813Z} &&&& 0.198 & 1.836 & 6.565 & 0.275 & 0.718 & 0.901 & 0.960 \\
        GeoNet \cite{yin2018geonet}  &&&& 0.153 & 1.328 & 5.737 & 0.232 & 0.802 & 0.934 & 0.972 \\
    \hline
        Ours (stereo only) &\checkmark & \checkmark && 0.090 & 0.844 & 4.373 & 0.190 & 0.900 & 0.954 & 0.976 \\
        Ours (flow + stereo) & \checkmark & \checkmark && 0.094 & 0.791 & 4.455 & 0.188 & 0.897 & 0.957 & 0.978 \\
        Ours (full-1) & \checkmark & \checkmark && 0.089 & 0.766 & 4.369 & \textbf{0.183} & 0.905 & \textbf{0.959} & \textbf{0.979} \\
        Ours (full-2) & \checkmark & \checkmark && 0.088 & \textbf{0.759} & \textbf{4.346} & 0.184 & \textbf{0.906} & \textbf{0.959} & \textbf{0.979} \\
        Ours (full-3) & \checkmark & \checkmark && \textbf{0.087} & 0.765 & 4.380 & 0.184 & \textbf{0.906} & \textbf{0.959} & 0.978 \\
    \hline
    \end{tabular}
    \label{tab:eigen}
    \vspace{-5mm}
\end{table*}
\section{Experimental Results}\label{sec:exp}
We evaluate the proposed method for both depth estimation and flow estimation on the KITTI dataset~\cite{Geiger2012CVPR}. We show that our framework is able to achieve competitive performance on both tasks.
Moreover, to show the merit of jointly learning the shared feature representations, we conduct experiments to validate that introducing stereo and flow pairs improves both performance.
We further enforce geometric constraints to construct the spatiotemporal correspondence in the stereo video and show that such constraint improves the performance via our newly proposed warping function.
%
%
\subsection{Implementation Details}
During training, we use a batch of size 2, each with two adjacent stereo pairs, i.e., 4 stereo pairs and 4 flow pairs. Images are scaled to the size of $512 \times 256$. Our model is based on Monodepth~\cite{monodepth17} using ResNet-50 as the encoder, with modifications of the last layer before output at each scale to generate 2-channel maps including horizontal and vertical correspondence maps.
The data augmentation follows Monodepth, containing left-right flipping, color augmentation of random gamma, brightness and color shifts, where each augmentation type has 50\% of chances to be selected.
Each color augmentation is sampled by uniform distributions in the ranges of [0.8, 1.2], [0.5, 2.0], [0.8, 1.2] respectively. We use Adam as our optimizer with default parameter settings. The learning rate is set as $10^{-4}$and we apply learning rate decay which is halved every 3 epochs for 5 times when training on full training set.
Our hyper-parameters $\{\alpha$, $\beta$, $\lambda_{sm}$, $\lambda_{lr}$, $\lambda_{2warp}\}$ are set to $\{$0.85, 10, 10, 0.5, 0.2$\}$. 
When only training on stereo pairs, $\lambda_{lr}$ would be 1 for balancing the proportion of stereo pairs in a batch.
Please note that we use a model variant trained without 2-warp consistency loss (i.e. denoted as Ours (flow+stereo) in Table~\ref{tab:kitti},~\ref{tab:eigen}, and \ref{tab:flow}) for better initializing the learning of our proposed full models. The source code and models are available on \url{ https://github.com/lelimite4444/BridgeDepthFlow}.

\subsection{Dataset and Setting}

The KITTI dataset contains stereo sequences of real road scenes, providing accurate but sparse depth and optical flow ground truth for a small subset. We evaluate our method on the KITTI 2012 and 2015 datasets, in which there are 194 and 200 pairs of flow and stereo with high quality annotations, covering 28 scenes of the KITTI raw dataset. During training, we generate 28968 cycles (i.e., a cycle contains 4 images as in Figure~\ref{fig:corr}) from the remaining 33 scenes.

To compare with other methods on depth estimation from the test set split by Eigen et al.~\cite{eigen2014depth}, which contains 697 pairs from 29 scenes in the KITTI raw dataset, We use the remaining 32 scenes and sample a subset consisting of 8000 cycles for training. We cap the depth to 0-80 meters with the same crop as in Garg \etal~\cite{garg2016unsupervised} during evaluation.

\subsection{Results for Depth Estimation}

\paragraph{KITTI Split.}\label{para:kitti}
In Table~\ref{tab:kitti}, we compare our results with state-of-the-art approaches~\cite{Wang_2018_CVPR,monodepth17,2018arXiv180610556Y} categorized by the use of stereo pairs during training and testing.
Compared to \cite{monodepth17} with the same setting, our models considering both the flow and stereo pairs consistently performs better in all the metric.
Note that, we use the same number of training data in training all the models for fair comparisons.
With comparing among our variants, adding flow pairs that are jointly learned within the same model with stereo pairs improves the base model (i.e., stereo only) by a significant margin. Further including the proposed 2-warp geometric connections brings additional gains in performance, using either type of the 2-warp operations as in Figure~\ref{fig:2warp}.

\paragraph{Eigen Split.}
In Table~\ref{tab:eigen}, we show depth estimation performance compared to state-of-the-art methods~\cite{eigen2014depth,monodepth17,2018arXiv180610556Y,2017arXiv170407813Z,yin2018geonet} on the Eigen split.
While existing methods do not have the same setting of using stereo pairs during training/testing, we show that our model significantly improves the performance by adding stereo pairs during testing, in which such data can be obtained as a free resource that only slightly increases the computational cost.
Note that, adding flow pairs without 2-warp consistency does not significantly improve the performance here in this Eigen split. The potential cause is due to the nature that flow estimation is considered to be harder than stereo matching, and the Eigen split is much smaller than the KITTI split. Therefore, learning optical flow simultaneously could lead to slower convergence and affect the performance of stereo matching. After we advance to include 2-warp consistency objective in our full models, it successfully overcomes the aforementioned issue and improves the performance, as now each stereo pair or temporally adjacent one can contribute multiple times to the same network $P$ by the proposed 2-warp function.

\begin{table*}[!t]
    \centering
    \caption{Quantitative evaluation on the optical flow task. EPE means average end-point-error where the postfix ``-noc'' and ``-occ'' only accounts for non-occlusion regions and occlusion regions, respectively. Fl means the error rate of the flow map values where one pixel is considered wrong if the EPE is $<$3px or $<$5\%.}
    \vspace{1mm}
    \scalebox{0.93}{
    \begin{tabular}{lcc|cc|cccccc}
    \hline 
        &&& \multicolumn{2}{c|}{KITTI 2012} & \multicolumn{6}{c}{KITTI 2015} \\
        Method & Train & Super- & train & train & train & train & train & train & train & train \\
        & Stereo & vised & EPE-all & EPE-noc & EPE-all & Fl-all & EPE-noc & Fl-noc & EPE-occ & Fl-occ \\
    \hline
        Flownet2 \cite{Ilg2017FlowNet2E} && \checkmark & 4.09 & - & 10.06 & 30.37\% & - & - & - & - \\
        Flownet2-CSS \cite{Ilg2017FlowNet2E} && \checkmark & 3.55 & - & 8.94 & 29.77\% & - & - & - & -\\
        PWC-Net \cite{Sun2018PWC-Net} && \checkmark & 4.14 && 10.35 & 33.67\% & - & - & - & -\\
        UnFlow-CSS \cite{Meister:2018:UUL} &&& 3.29 & \textbf{1.26} & 8.10 & \textbf{23.27\%} & - & - & - & -\\
        GeoNet \cite{yin2018geonet} &&& - & - & 10.81 & - & 8.05 & - & - & -\\
        DF-net \cite{zou2018dfnet} &&& 3.54 & - & 8.98 & 26.01\% & - & - & - & -\\
    \hline
        Ours (flow only) &&& 4.29 & 1.98 & 9.70 & 32.77\% & 5.23 & 25.89\% & 26.06 & 65.08\%\\
        Ours (flow + stereo) & \checkmark && 2.64 & 1.45 & 7.47 & 28.54\% & 4.707 & 22.56\% & 17.83 & 56.29\%\\
       Ours (w/o sharing) & \checkmark && 3.49 & 1.99 & 8.78 & 34.56\% & 5.33 & 28.65\% & 21.38 & 62.61\%\\
        Ours (full-1) & \checkmark && 2.59 & 1.41 & \textbf{7.021} & 27.34\% & 4.257 & 21.41\% & \textbf{17.57} & 54.78\% \\
        Ours (full-2) & \checkmark && 2.61 & 1.39 & 7.044 & 27.73\% & \textbf{4.229} & 21.65\% & 17.89 & 55.74\% \\
        Ours (full-3) & \checkmark && \textbf{2.56} & 1.388 & 7.134 & 27.13\% & 4.306 & \textbf{21.19\%} & 17.79 & \textbf{54.09\%}\\
    \hline
    \end{tabular}}
    \label{tab:flow}
    \vspace{5mm}
\end{table*}

\begin{figure*}
\begin{center}
\begin{tabular}{c@{\hskip 1mm}c@{\hskip 1mm}c@{\hskip 1mm}c@{\hskip 1mm}c}
    Image & Our Depth Map & GT Depth Map & Our Flow Map & GT Flow Map\\
      \vspace{0mm}
      {\hspace{0mm}}
    \includegraphics[height=0.1\linewidth,width=0.19\linewidth]{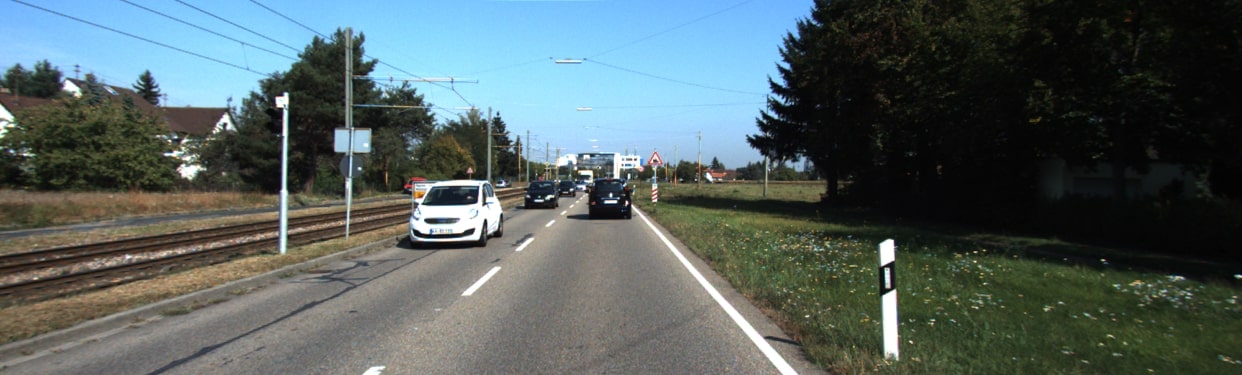}&\includegraphics[height=0.1\linewidth,width=0.19\linewidth]{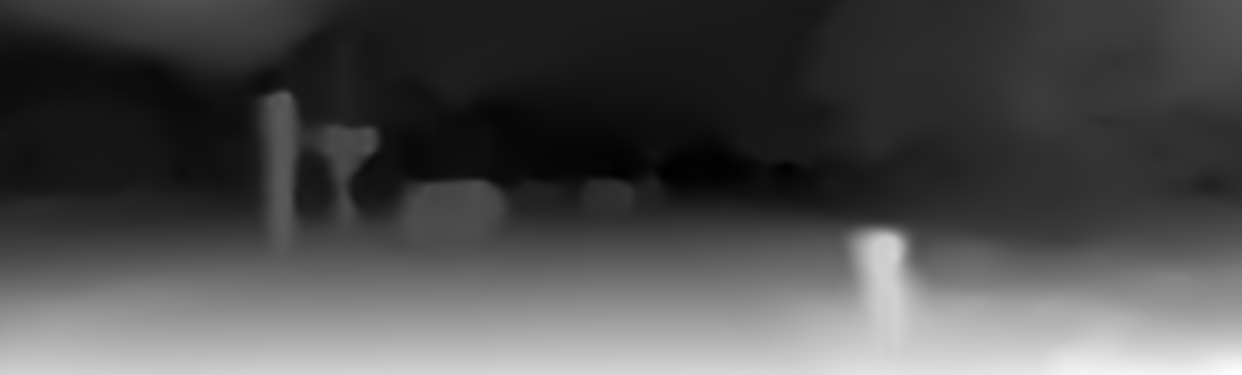}&\includegraphics[height=0.1\linewidth,width=0.19\linewidth]{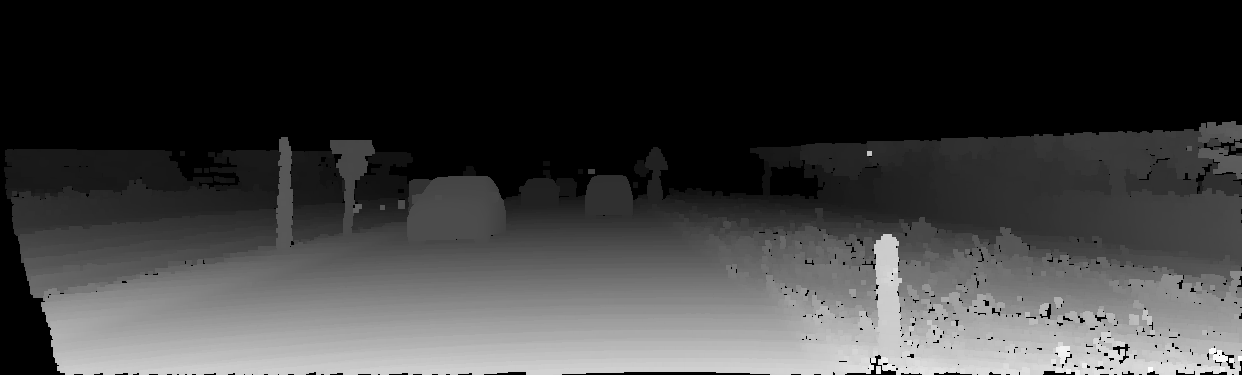}&\includegraphics[height=0.1\linewidth,width=0.19\linewidth]{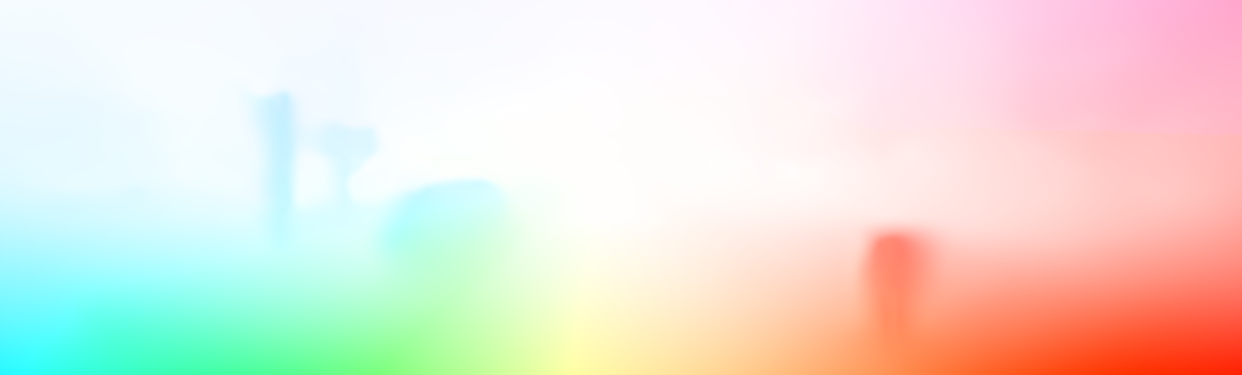}&\includegraphics[height=0.1\linewidth,width=0.19\linewidth]{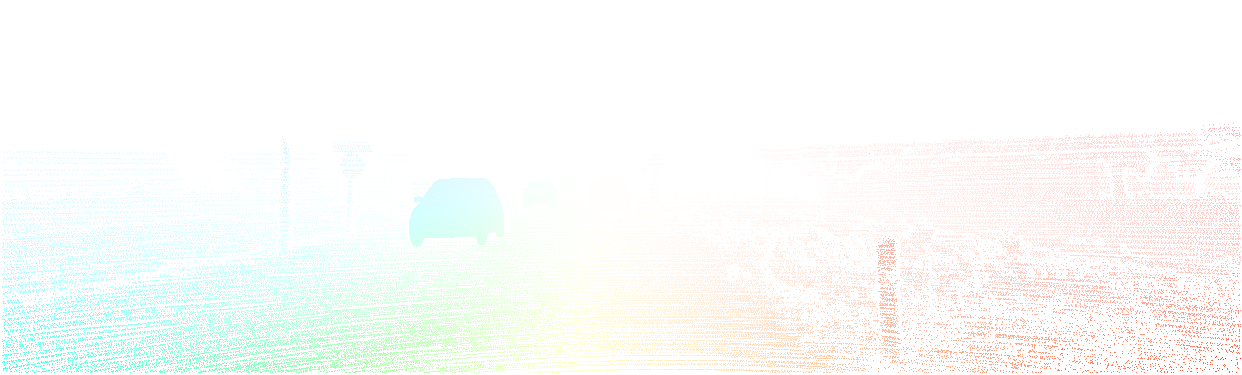}\\
      \vspace{0mm}
      {\hspace{0mm}}
    \includegraphics[height=0.1\linewidth,width=0.19\linewidth]{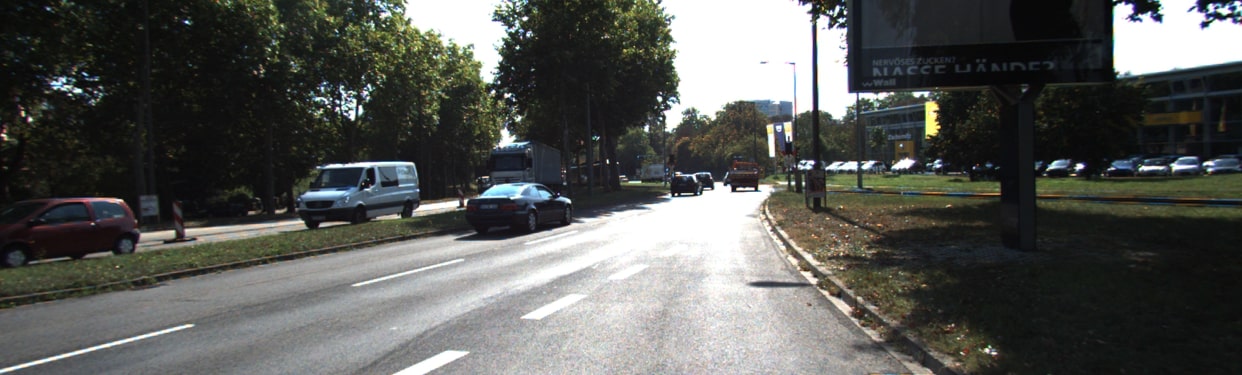}&\includegraphics[height=0.1\linewidth,width=0.19\linewidth]{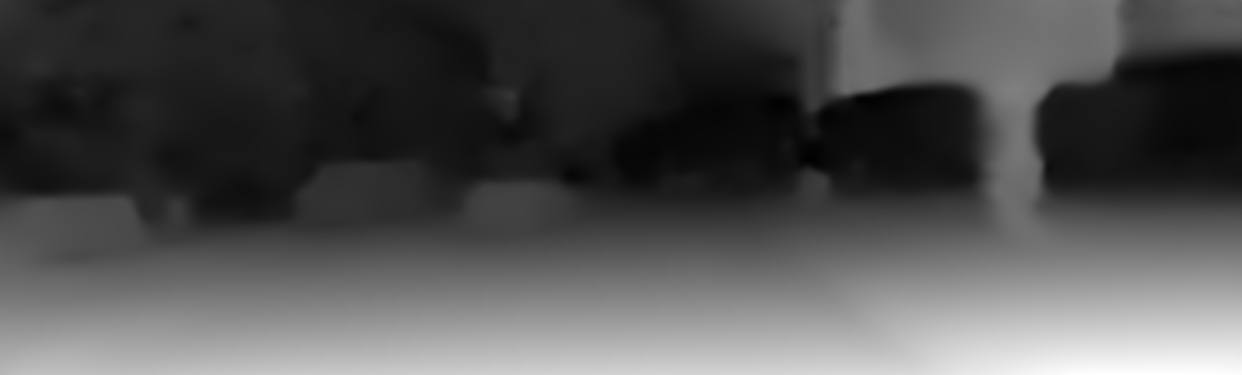}&\includegraphics[height=0.1\linewidth,width=0.19\linewidth]{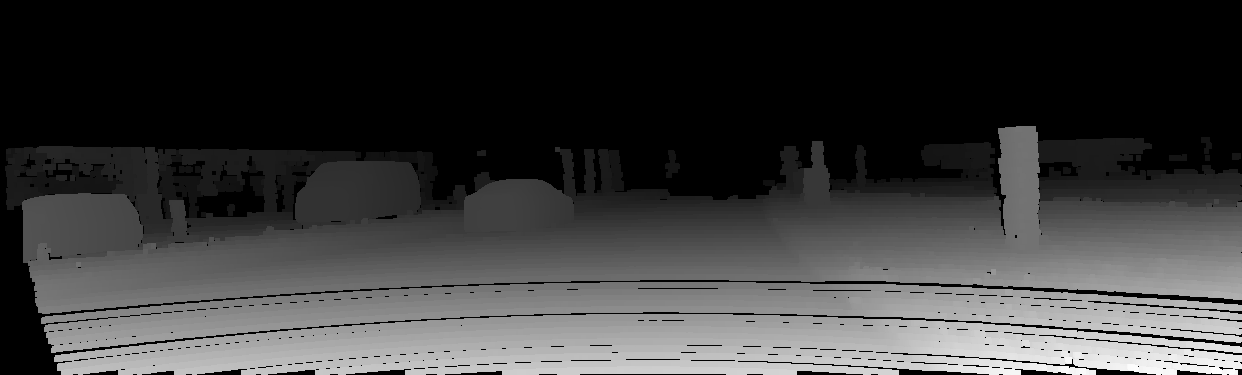}&\includegraphics[height=0.1\linewidth,width=0.19\linewidth]{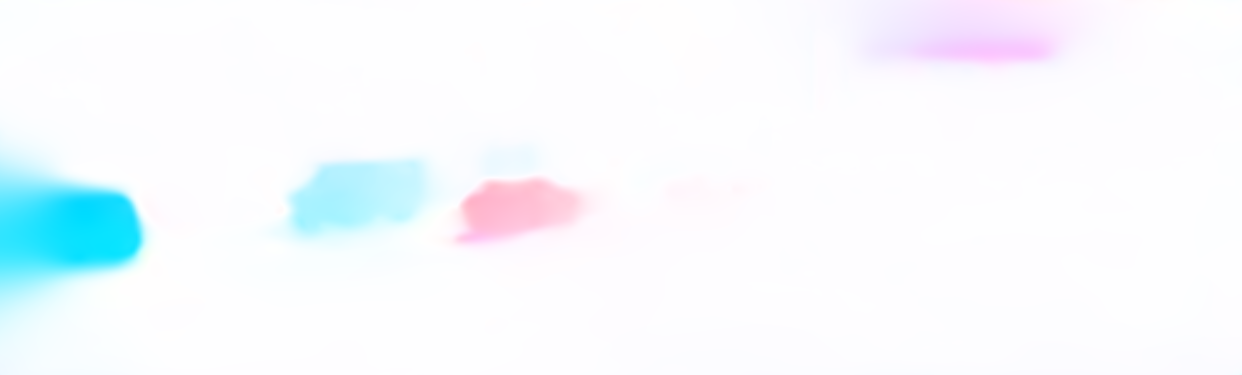}&\includegraphics[height=0.1\linewidth,width=0.19\linewidth]{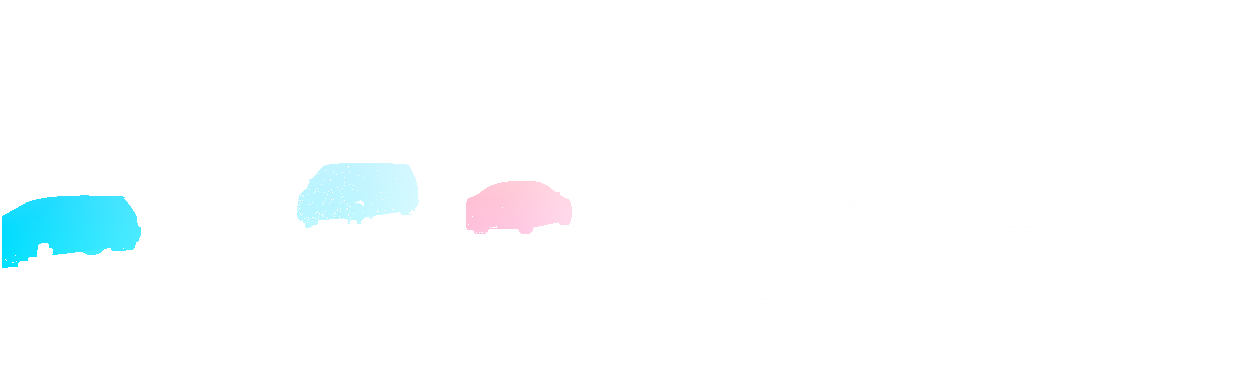}\\
      \vspace{0mm}
      {\hspace{0mm}}
    \includegraphics[height=0.1\linewidth,width=0.19\linewidth]{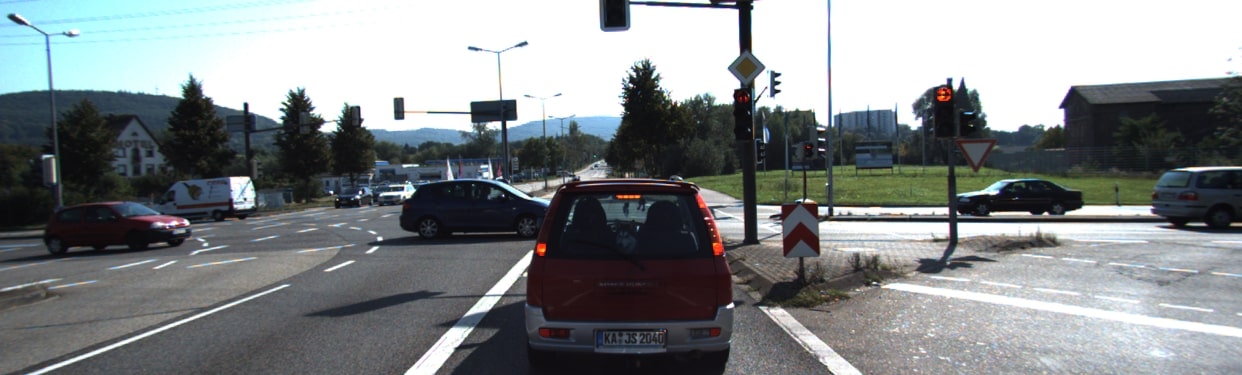}&\includegraphics[height=0.1\linewidth,width=0.19\linewidth]{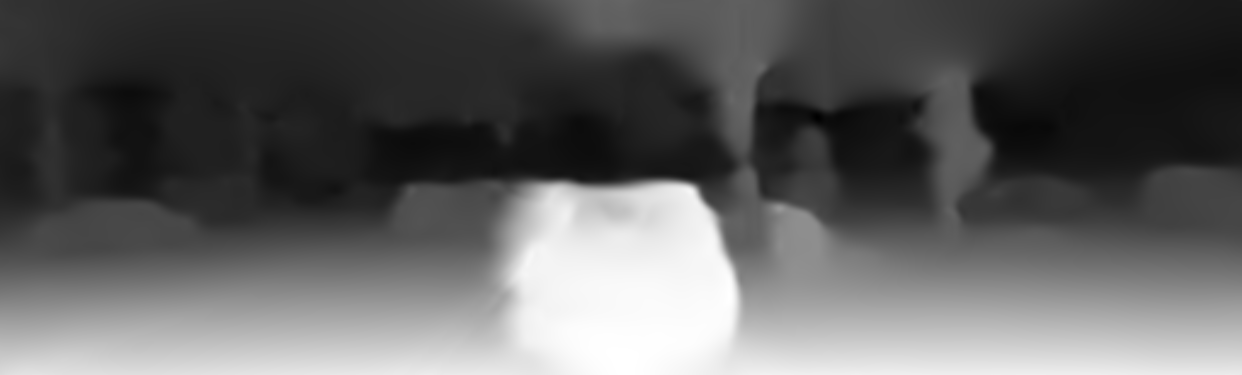}&\includegraphics[height=0.1\linewidth,width=0.19\linewidth]{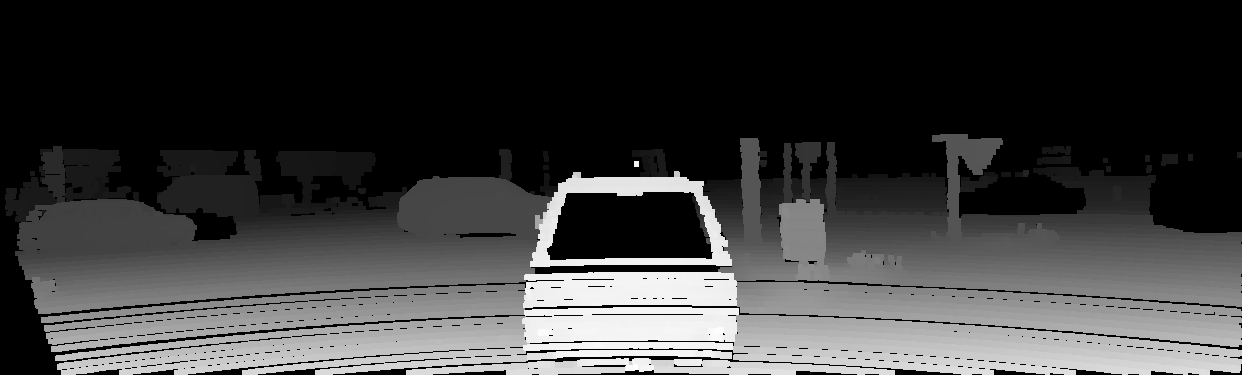}&\includegraphics[height=0.1\linewidth,width=0.19\linewidth]{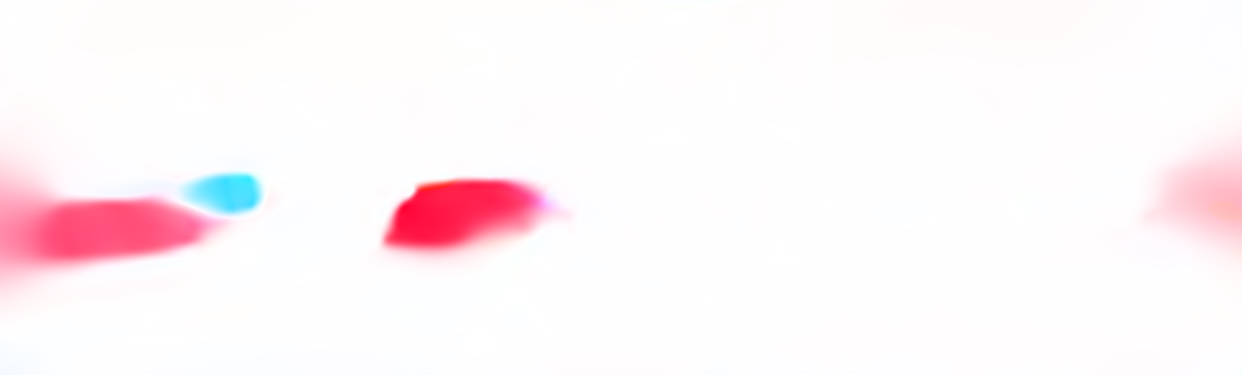}&\includegraphics[height=0.1\linewidth,width=0.19\linewidth]{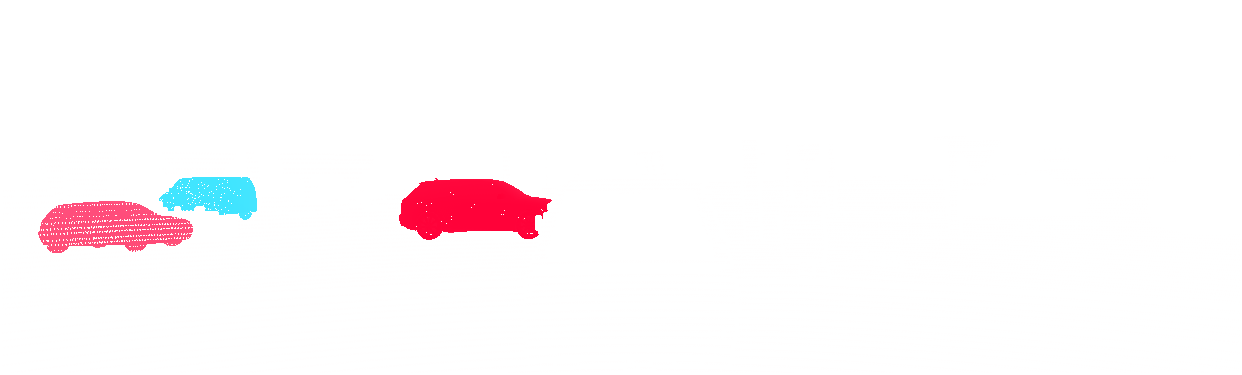}\\
      \vspace{0mm}
      {\hspace{0mm}}
    \includegraphics[height=0.1\linewidth,width=0.19\linewidth]{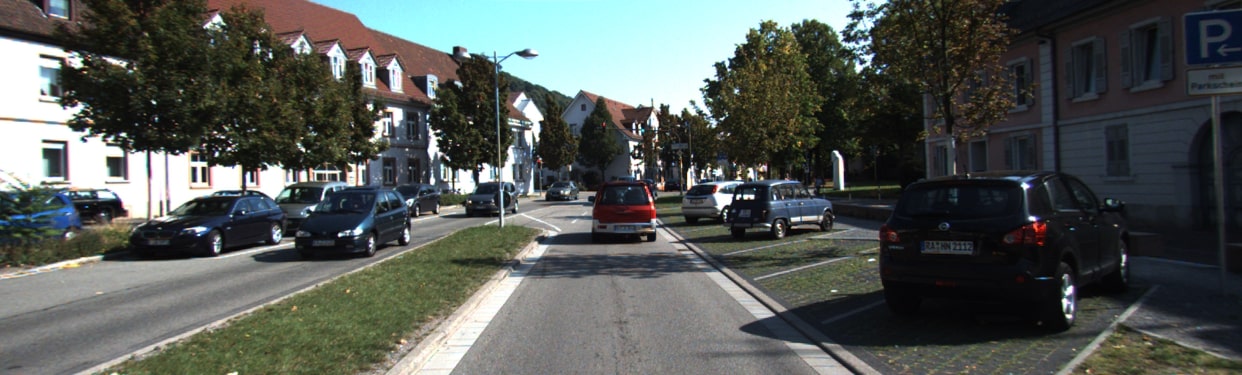}&\includegraphics[height=0.1\linewidth,width=0.19\linewidth]{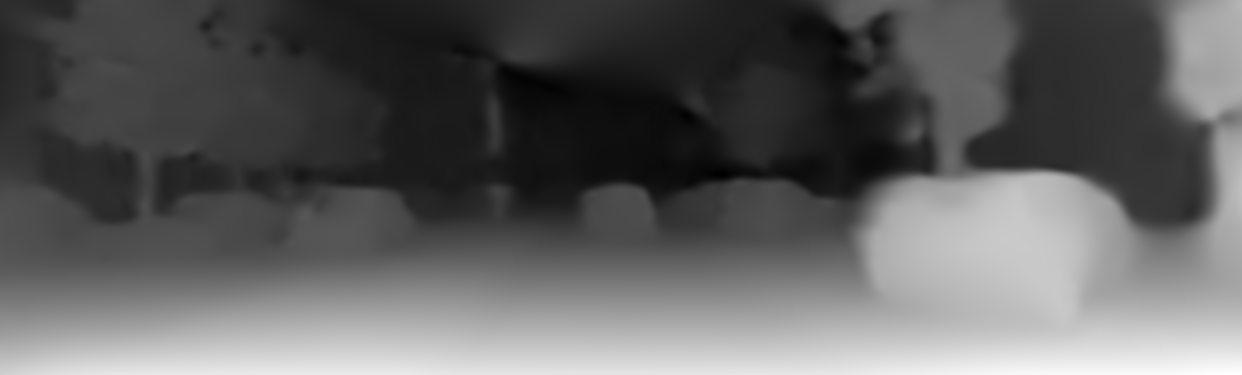}&\includegraphics[height=0.1\linewidth,width=0.19\linewidth]{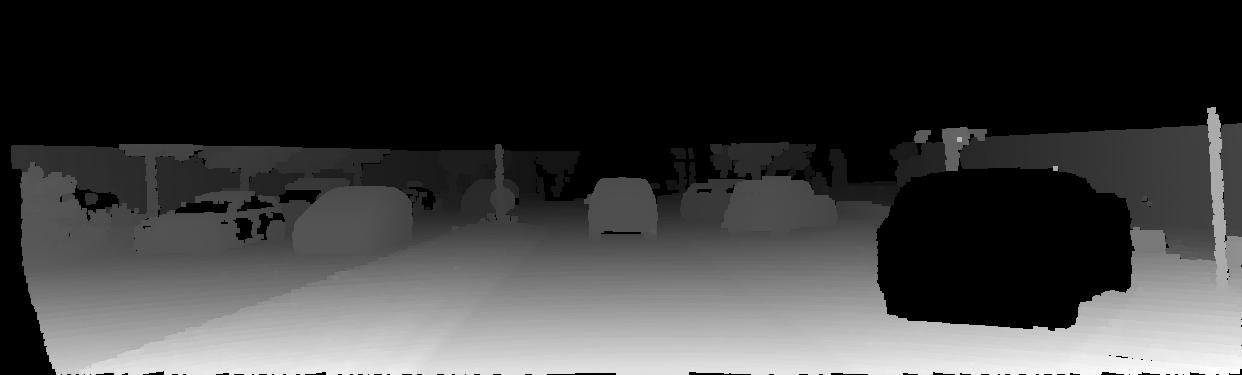}&\includegraphics[height=0.1\linewidth,width=0.19\linewidth]{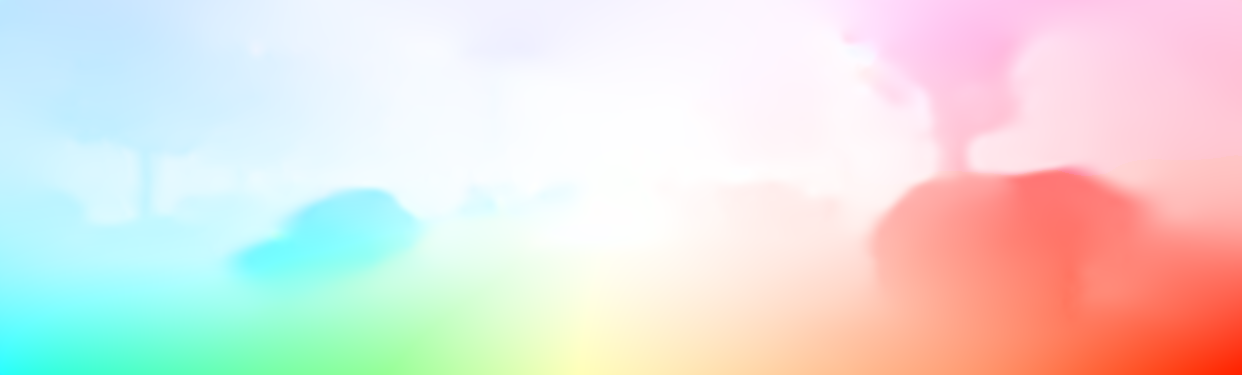}&\includegraphics[height=0.1\linewidth,width=0.19\linewidth]{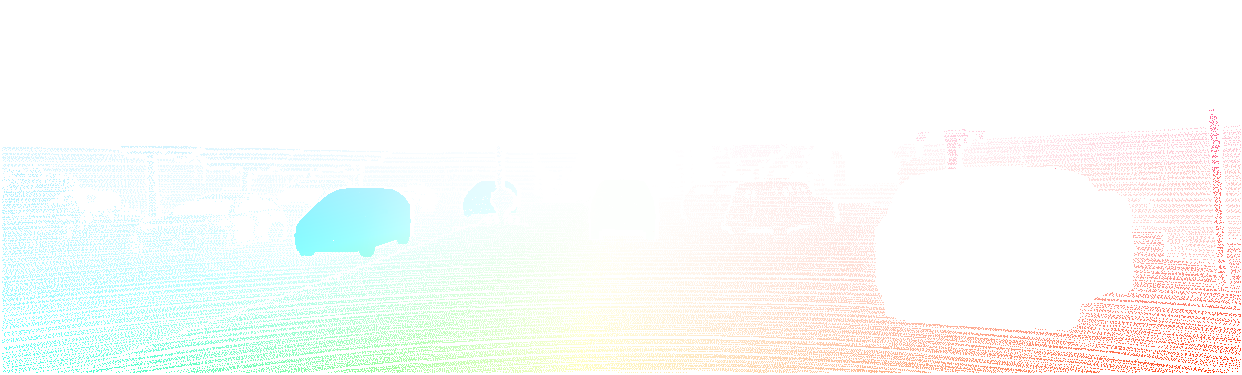}\\
\end{tabular}
\end{center}
\caption{Example results on KITTI. In each row, we sequentially show the left image at time $t$, our predicted depth map, the ground truth depth, our flow prediction, and the ground truth flow.}
\label{fig:comparison}
\end{figure*}
\subsection{Results for Flow Estimation}
In Table~\ref{tab:flow}, we show our unsupervised flow results compared with state-of-the-art supervised methods~\cite{Ilg2017FlowNet2E,Sun2018PWC-Net} and unsupervised approaches~\cite{Meister:2018:UUL,yin2018geonet,zou2018dfnet}.
The results suggest that our model without using 2-warp already performs favorably against other unsupervised framework.
It demonstrates the benefit of using a single network to jointly learn feature representations shared across two highly co-related tasks (i.e., flow estimation and stereo matching) and help improve both performance.
In addition, even optical flow estimation is a harder task, our proposed 2-warp consistency loss is able to encourage the tighter connection across two tasks and thus further boost the performance. We also observe from the KITTI 2015 dataset that all three variants of our full model achieve similar improvement in a significant margin, in which it shows that our proposed 2-warp consistency loss could benefit the estimation of pixel correspondences regardless the warping directions.
\subsection{Results without sharing weights}

In order to demonstrate the benefit of using a single network for both flow estimation and stereo matching instead of having separate architectures for each task, we train a model variant of full-2 with untying the weights of both tasks and test it on KITTI 2015, which is denoted as Ours (w/o sharing) in Table~\ref{tab:flow}. 
We find its performance comparable to our full model in stereo matching but much worse in optical flow estimation. 
The main reason is that the learning rates for flow and depth networks are now hard to balance without well-tuning, and the performance of optical flow estimation becomes unstable for the 2-warp operation. This shows the advantage of having a single and principled network for both tasks.
We show some example results in Figure~\ref{fig:comparison}.
\section{Conclusions}\label{sec:conclusion}
In this paper, we propose to use a single, principled network to perform both stereo matching and flow estimation.
The advantage lies in that the feature representations can be jointly learned and shared across two tasks, which all aim to predict pixel correspondences, spatially and temporally.
Given a stereo video, we further enforce geometric connections between adjacent stereo pairs, in which a 2-warp consistency term is introduced to optimize the reconstruction loss via the warping functions.
Experimental results show that the proposed framework facilitates the information from two tasks and thus improves the performance on both depth and flow estimation.

\noindent\textbf{\large Acknowledgement} This project is supported by MOST-108-2636-E-009-001 and we are grateful to the National Center for High-performance Computing, Taiwan, for computer time and facilities, as well as the support of NVIDIA Corporation with the donation of the Titan Xp GPU used for this research.

{\small
\bibliographystyle{ieee}
\bibliography{egbib}
}
\clearpage
\begin{figure*}[b]
\centering
  \includegraphics[width=0.95\linewidth]{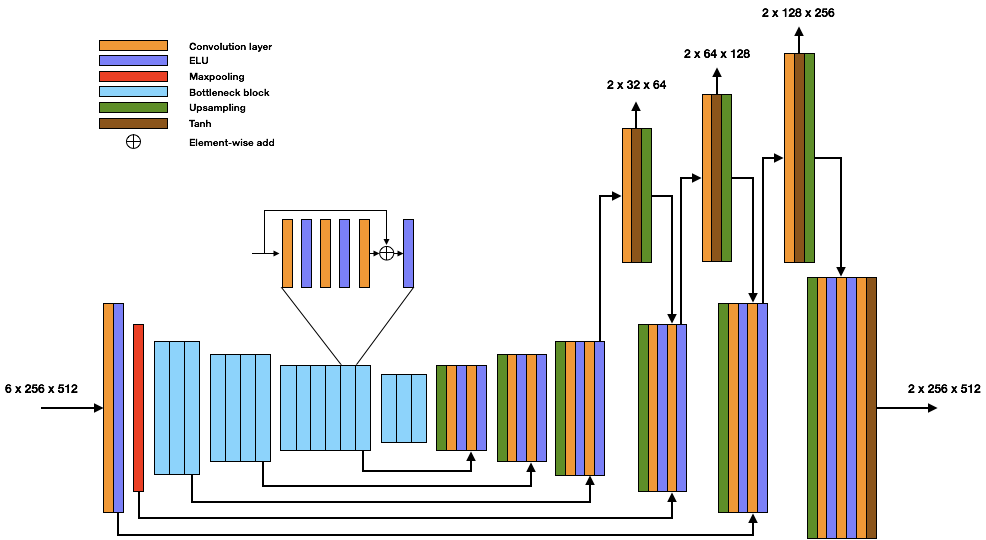}
  \caption{The detailed structure of our model. We use ResNet-50 as the encoder, where there are four bottleneck blocks with different spatial size (denoted as blue blocks). The last convolution layer of each block has the spatial size with stride equal to 2. In addition, each bottleneck block has its own skip connection with the element-wise addition before the last activation function. The decoder outputs four estimated maps at different scales and up-samples them as the input size toward the next decoder block. 
  }
  \label{fig:structure}
\end{figure*}

\section*{Network Architecture}
Figure \ref{fig:structure} shows the architecture of the proposed single model for both stereo matching and flow estimation. We concatenate an image pair as the input no matter they are stereo pairs or flow pairs on the color channel. The output is the correspondence map at 4 different scales and each is a 2-channel map indicating vertical and horizontal correspondence.

\section*{PWC-Net}
We utilize a more powerful architecture for optical flow estimation, i.e., PWC-Net, and adopt it in our proposed method. As shown in Table \ref{tab:pwc_stereo} and Table \ref{tab:pwc_flow}, the performance on both tasks improve significantly comparing to the Monodepth backbone (i.e., the \textit{full-2} model in the main paper).
For optical flow, our method is also competitive with Janai et al.~\cite{Janai2018ECCV}, which also uses PWC-Net in their proposed model.

\begin{table*}[!h]
    \centering
    \renewcommand{\arraystretch}{1.2}
    \setlength{\tabcolsep}{2pt}
    \caption{Quantitative evaluation on depth estimation using PWC-Net.}
    \vspace{1mm}
    \begin{tabular}{lcc|lccc|ccc}
    \hline 
        Method & Train & Test & \multicolumn{4}{c|}{Lower the better} & \multicolumn{3}{c}{Higher the better} \\
        & Stereo & Stereo & Abs Rel & Sq Rel & RMSE & RMSE log & $\delta < 1.25$ & $\delta < 1.25^2$ & $\delta < 1.25^3$ \\
    \hline
        Ours (Monodepth) & \checkmark & \checkmark & 0.088 & 0.759 & 4.346 & 0.184 & 0.906 & 0.959 & 0.979 \\
        Ours (PWC-Net) & \checkmark & \checkmark & 0.058 & 0.694 & 4.020 & 0.152 & 0.952 & 0.979 & 0.990 \\
    \hline
    \end{tabular}
    \label{tab:pwc_stereo}
\end{table*}

\begin{table*}[!t]
    \centering
    \caption{Quantitative evaluation on the optical flow task using PWC-Net.}
    \vspace{1mm}
    \scalebox{0.93}{
    \begin{tabular}{lcc|cc}
    \hline 
        &&& \multicolumn{2}{c}{KITTI 2015} \\
        Method & Stereo & Multi-frame & train & train \\
        & & & EPE-all & Fl-all \\
    \hline
        Ours (Monodepth) & \checkmark & & 7.04 & 27.73\% \\
        Ours (PWC-Net) & \checkmark & & 6.66 & 21.50\% \\
        Janai et al.~\cite{Janai2018ECCV} & & \checkmark & 6.59 & - \\
    \hline
    \end{tabular}}
    \label{tab:pwc_flow}
    \vspace{5mm}
\end{table*}

\section*{More Visual Results}
\begin{figure*}[!b]
\begin{tabular}{c@{\hskip 1mm}c@{\hskip 1mm}c}
    {\hspace{0mm}}
    \includegraphics[width=0.32\linewidth]{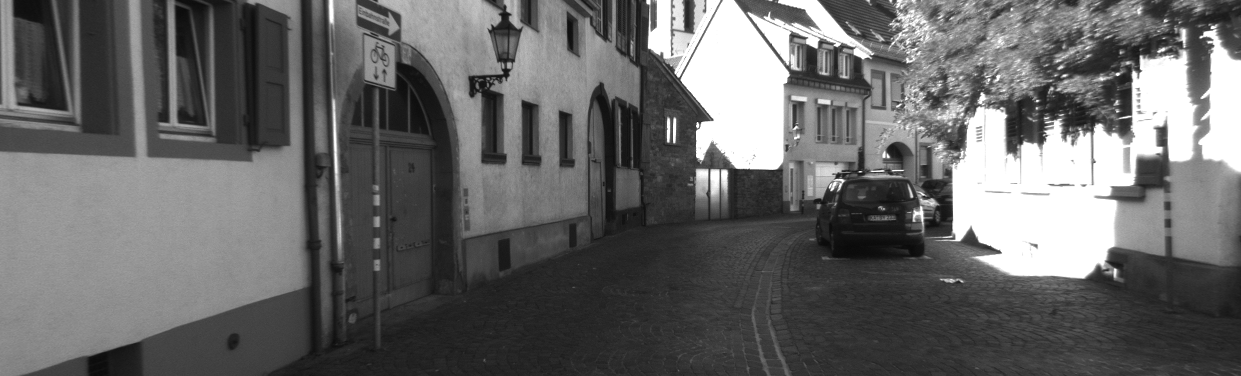}&\includegraphics[width=0.32\linewidth]{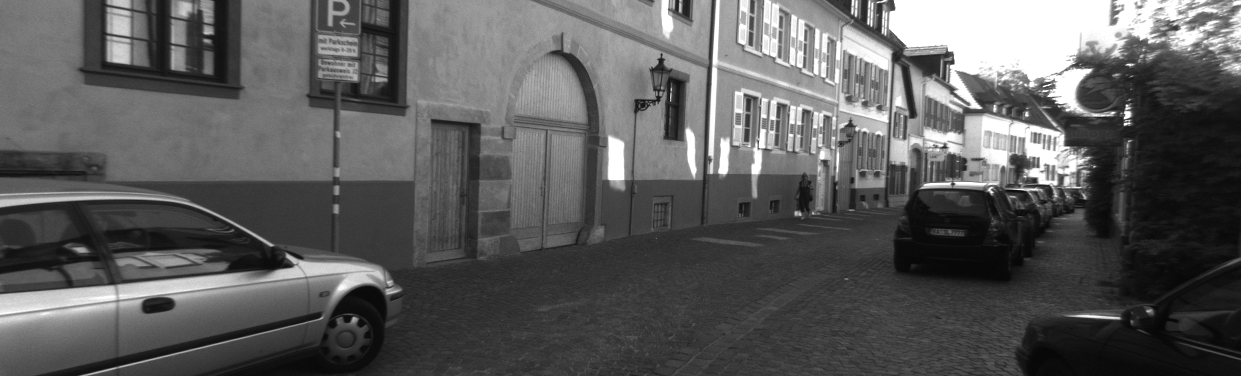}&\includegraphics[width=0.32\linewidth]{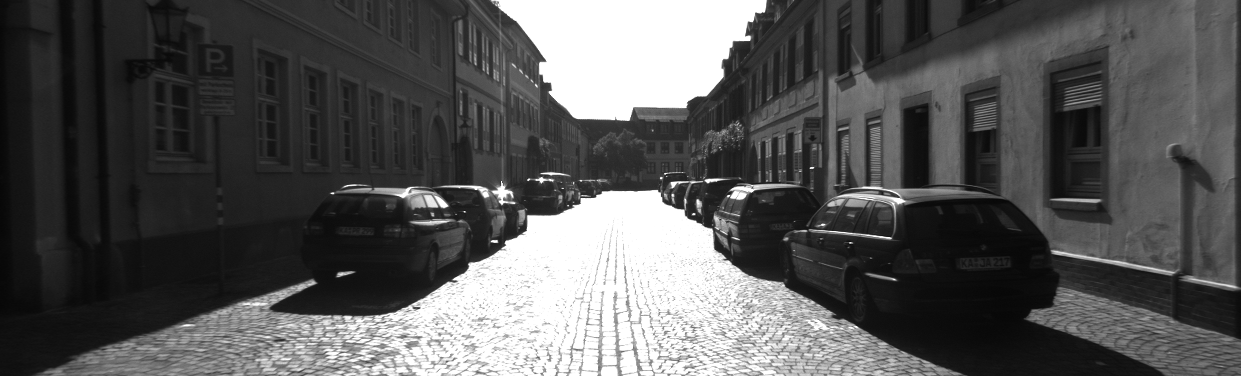}\\
    {\hspace{0mm}}
    \includegraphics[width=0.32\linewidth]{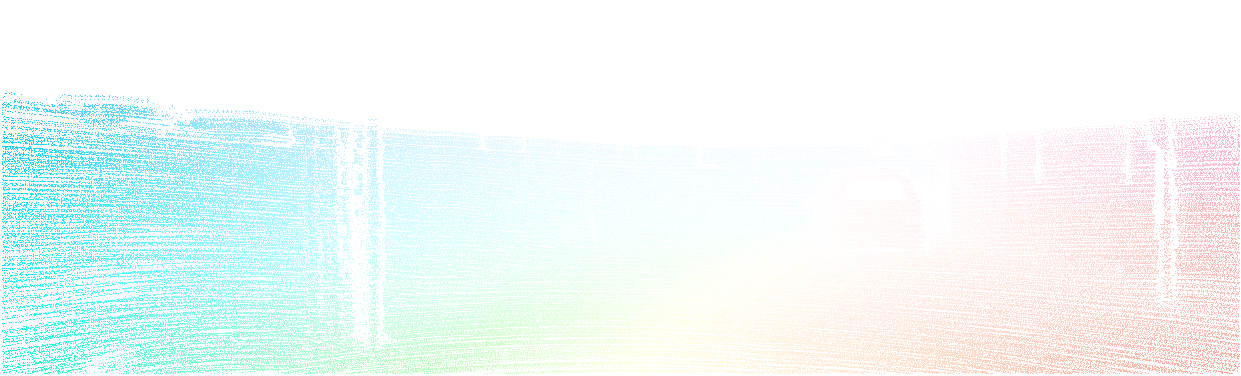}&\includegraphics[width=0.32\linewidth]{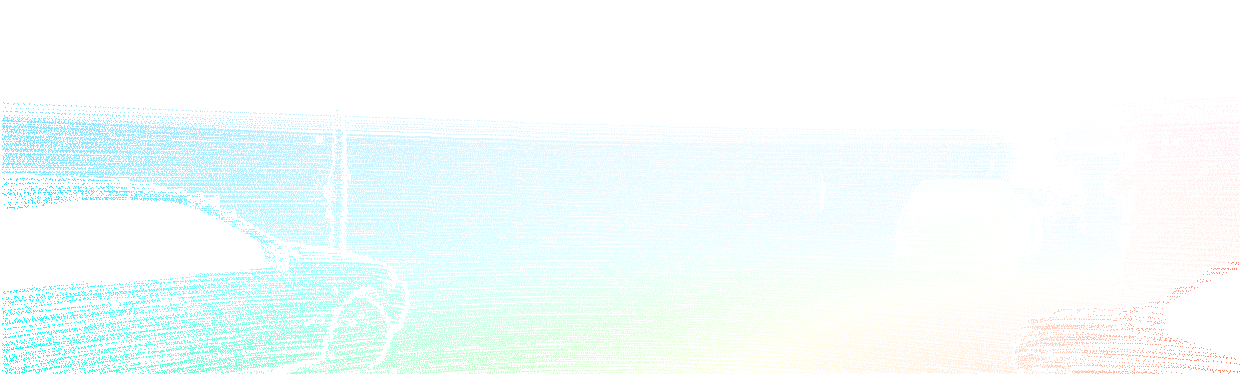}&\includegraphics[width=0.32\linewidth]{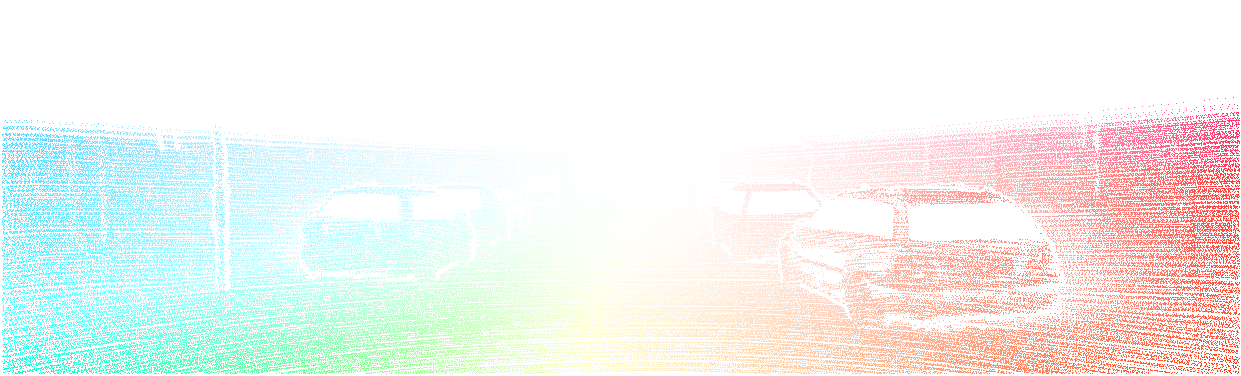}\\
    \vspace{3mm}
    {\hspace{0mm}}
    \includegraphics[width=0.32\linewidth]{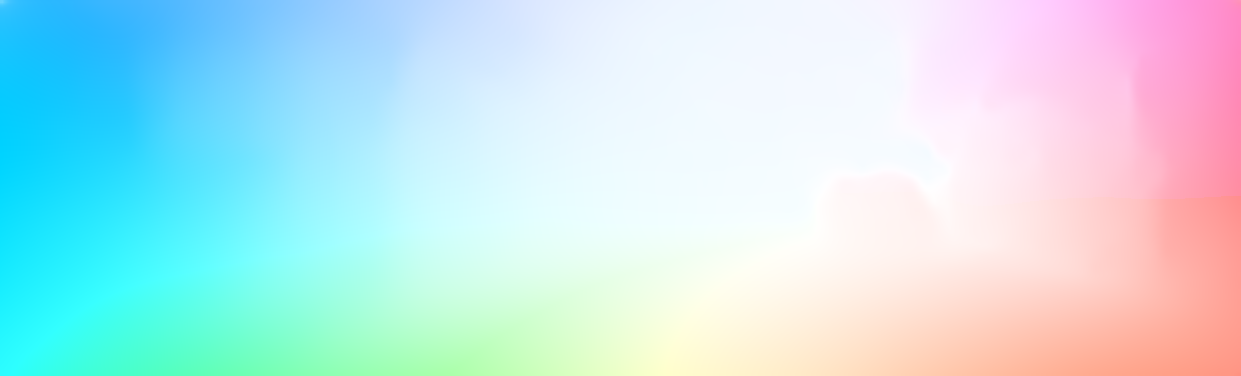}&\includegraphics[width=0.32\linewidth]{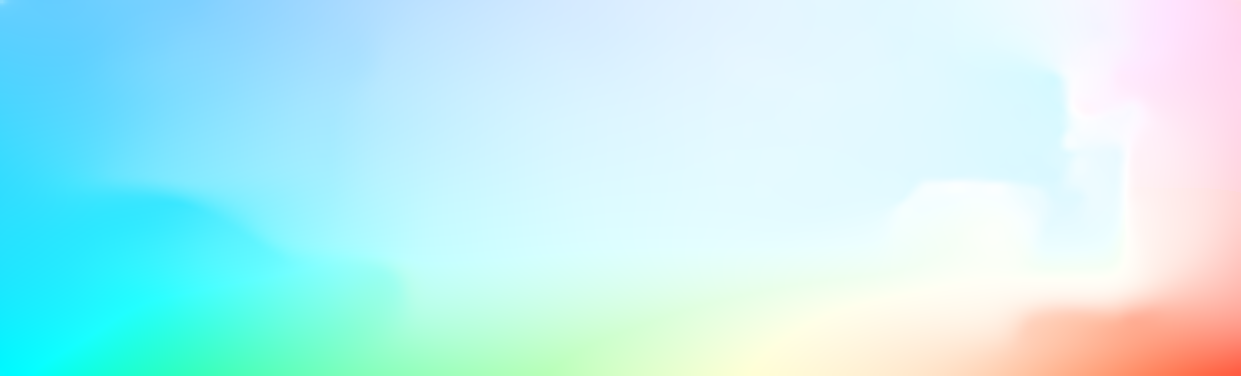}&\includegraphics[width=0.32\linewidth]{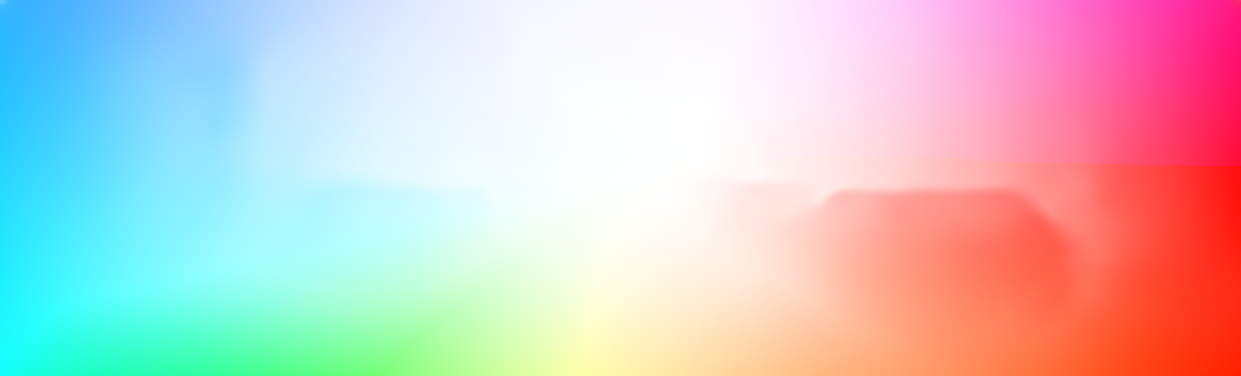}\\
    {\hspace{0mm}}
    \includegraphics[width=0.32\linewidth]{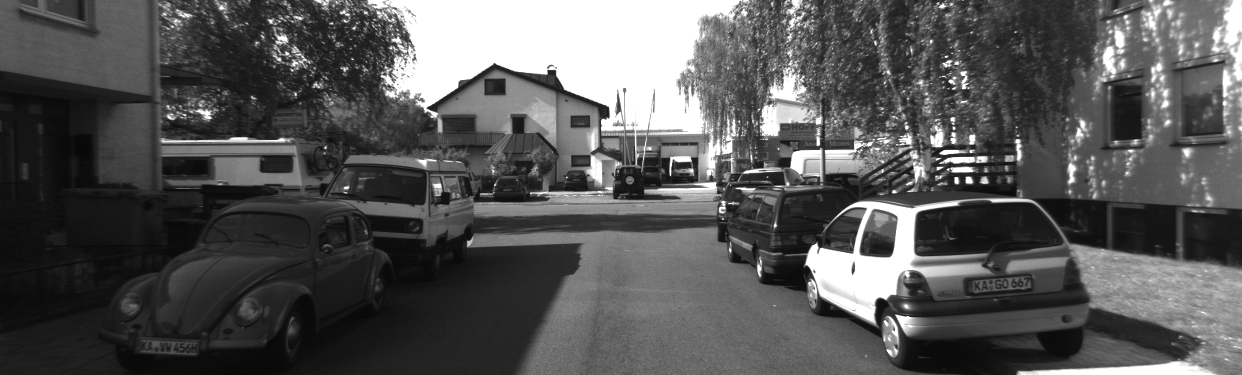}&\includegraphics[width=0.32\linewidth]{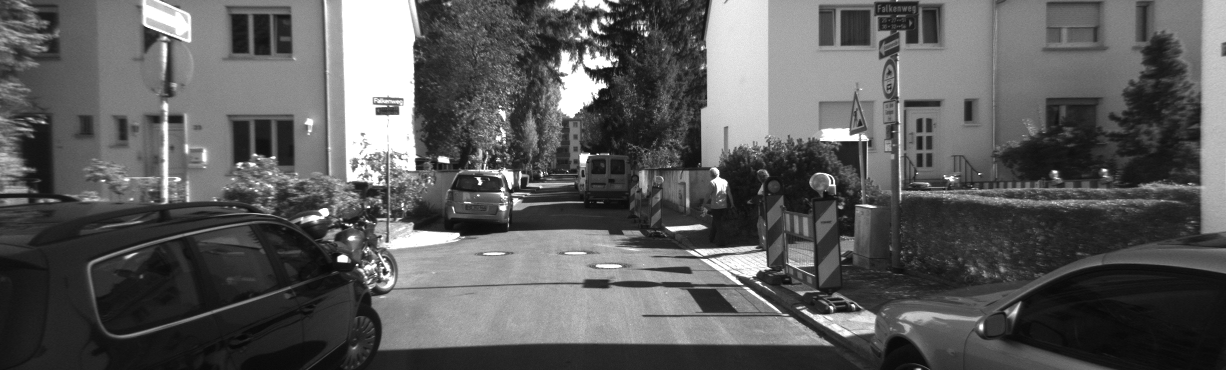}&\includegraphics[width=0.32\linewidth]{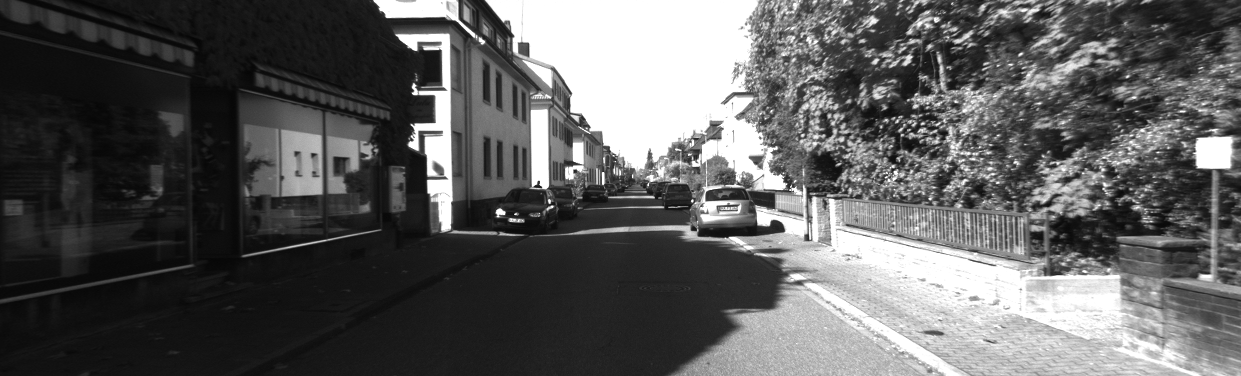}\\
    {\hspace{0mm}}
    \includegraphics[width=0.32\linewidth]{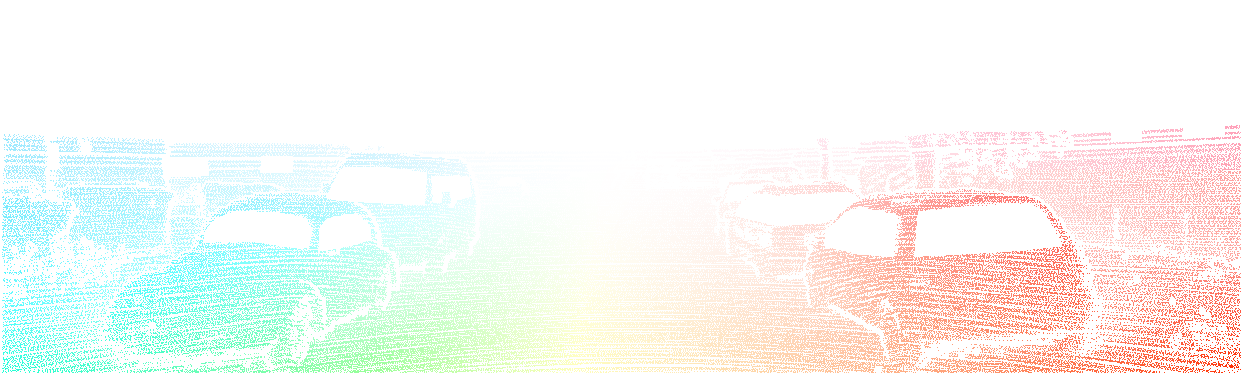}&\includegraphics[width=0.32\linewidth]{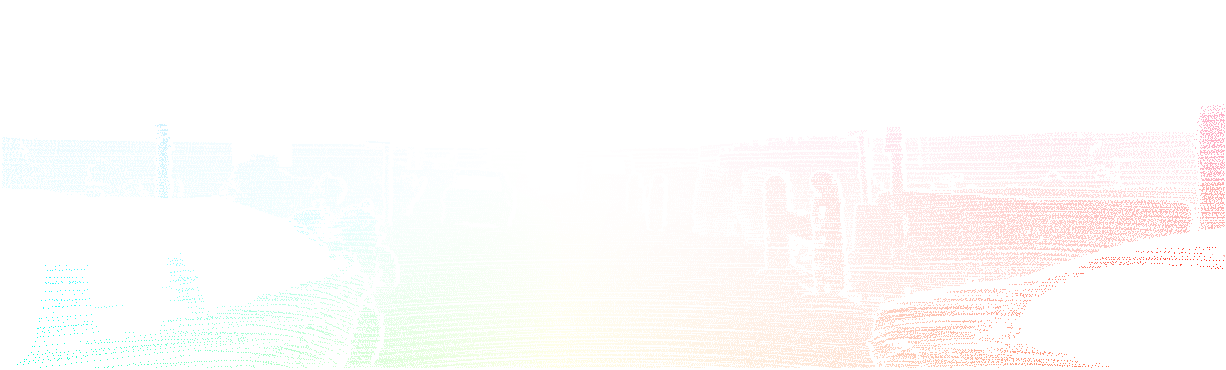}&\includegraphics[width=0.32\linewidth]{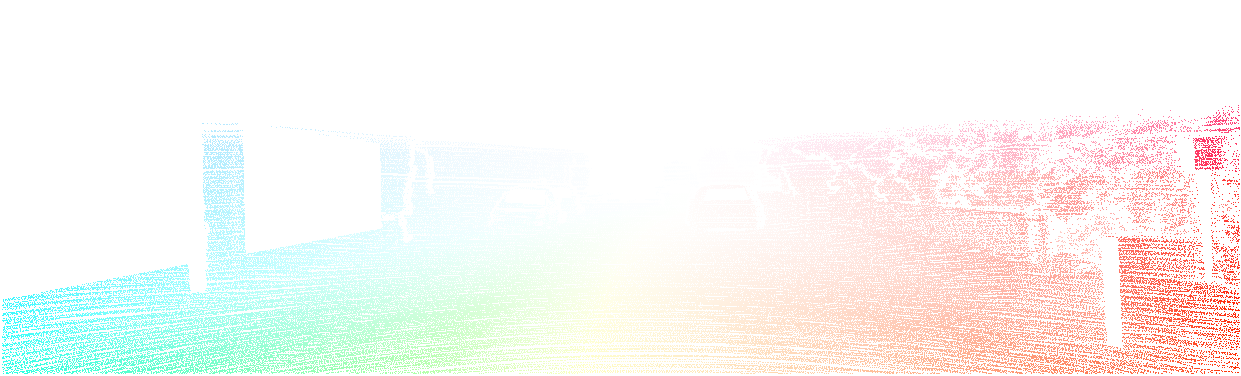}\\
    {\hspace{0mm}}
    \includegraphics[width=0.32\linewidth]{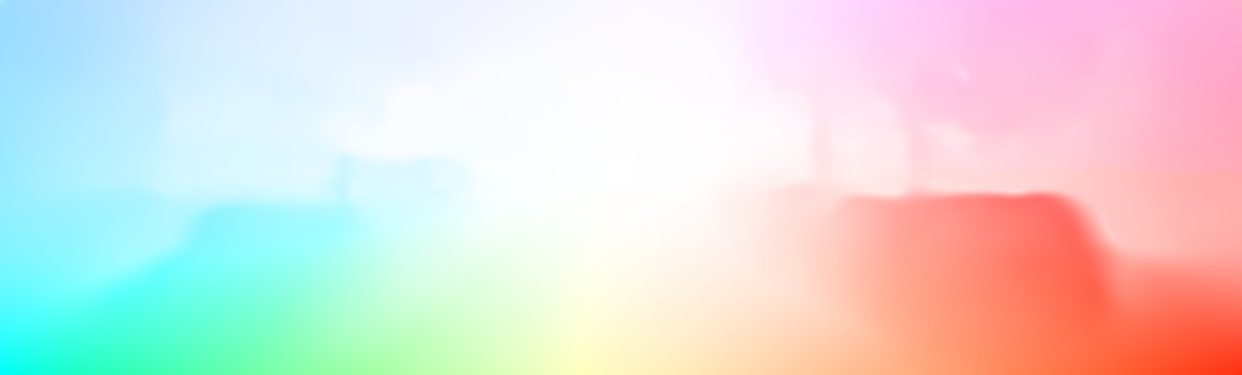}&\includegraphics[width=0.32\linewidth]{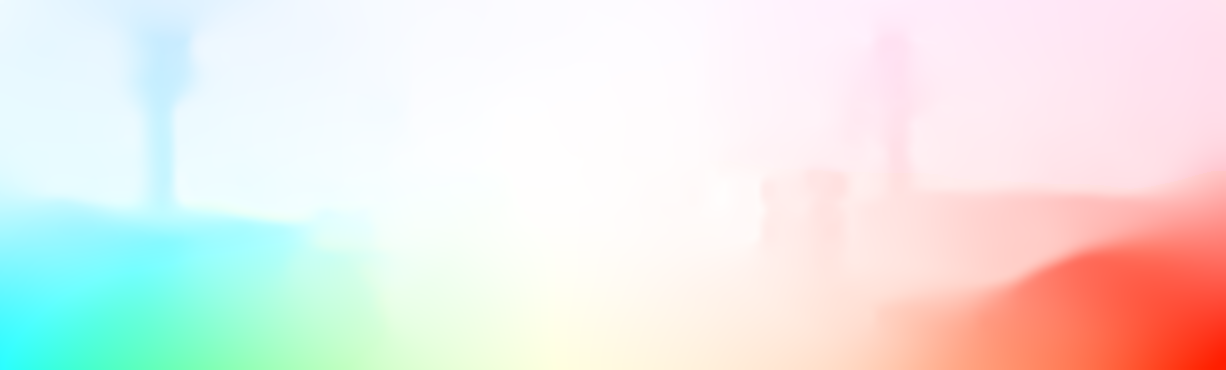}&\includegraphics[width=0.32\linewidth]{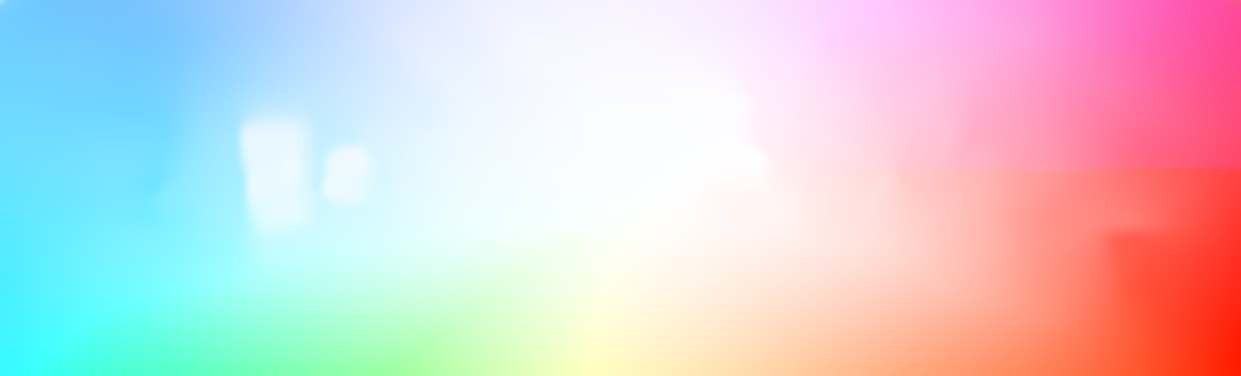}\\
\end{tabular}
\caption{Estimated optical flow results on KITTI 2012. There are 6 examples and from top to bottom, they are real gray-scale images, ground truths and our estimated results.}
\label{fig:2012}
\end{figure*}

Figure \ref{fig:2012} shows our optical flow results on the KITTI 2012 dataset which consists of rigid scenes, where the sparse ground truths are provided as the reference.
Figure \ref{fig:comparison} shows the depth maps of different variants of our models trained on only KITTI stereo pairs, both KITTI stereo and flow pairs, and our full model with the 2-warp consistency, respectively. Our full model can deal with challenging cases such as thin/small objects and have better response on the object boundaries. 

\begin{figure*}[!hb]
\centering
\begin{tabular}{c@{\hskip 1mm}c@{\hskip 1mm}c@{\hskip 1mm}c}
    Image & Stereo & Stereo + Flow & Full Model \\
      \vspace{3mm}
      {\hspace{0mm}}
    \includegraphics[width=0.23\linewidth]{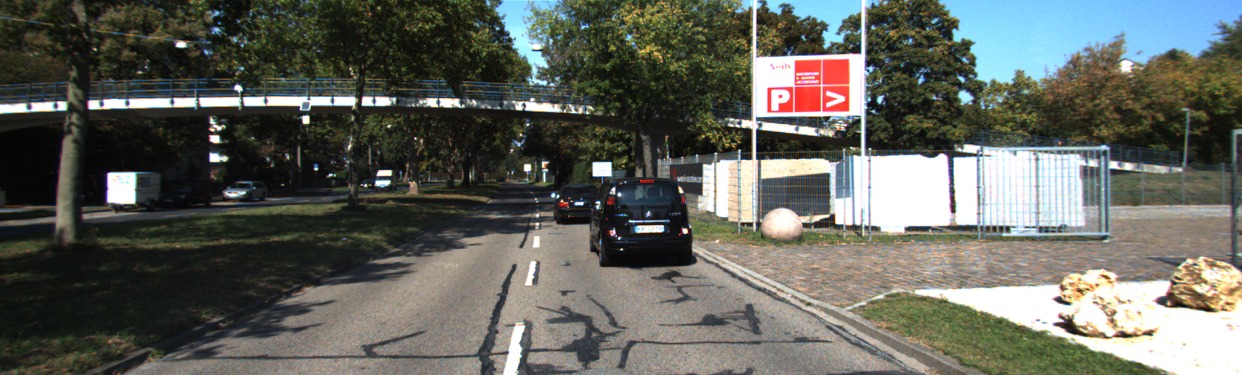}&\includegraphics[width=0.23\linewidth]{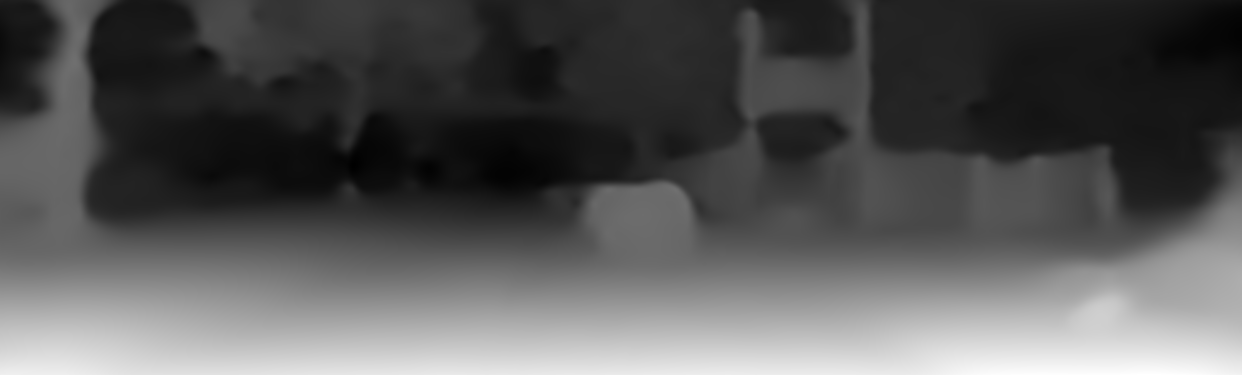}&\includegraphics[width=0.23\linewidth]{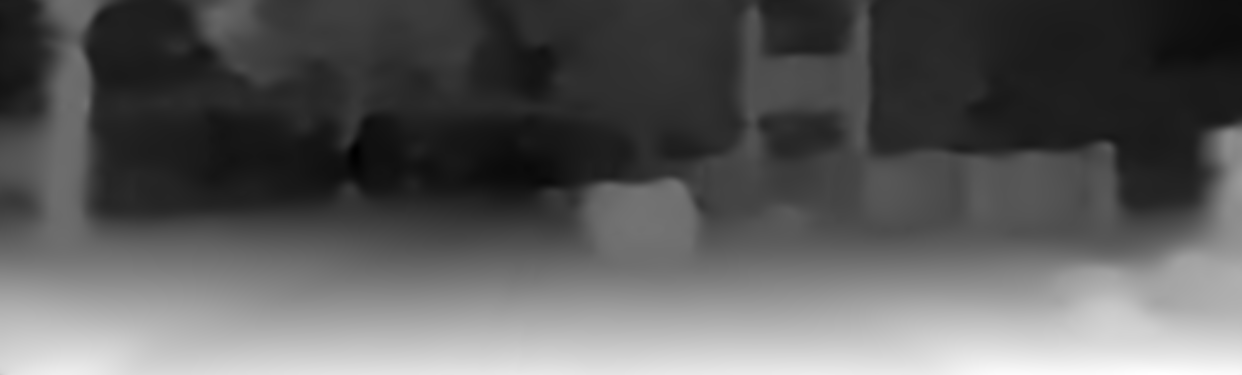}&\includegraphics[width=0.23\linewidth]{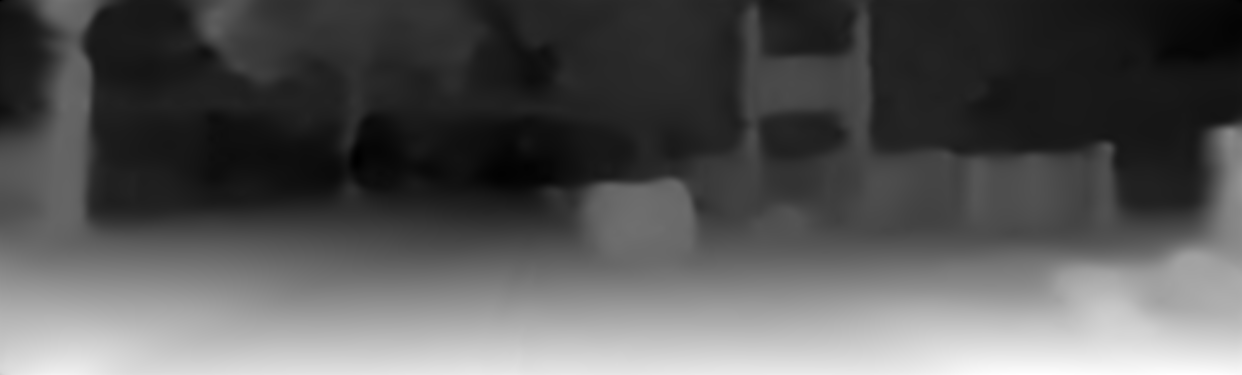}\\
    \vspace{3mm}
    {\hspace{0mm}}
    \includegraphics[width=0.23\linewidth]{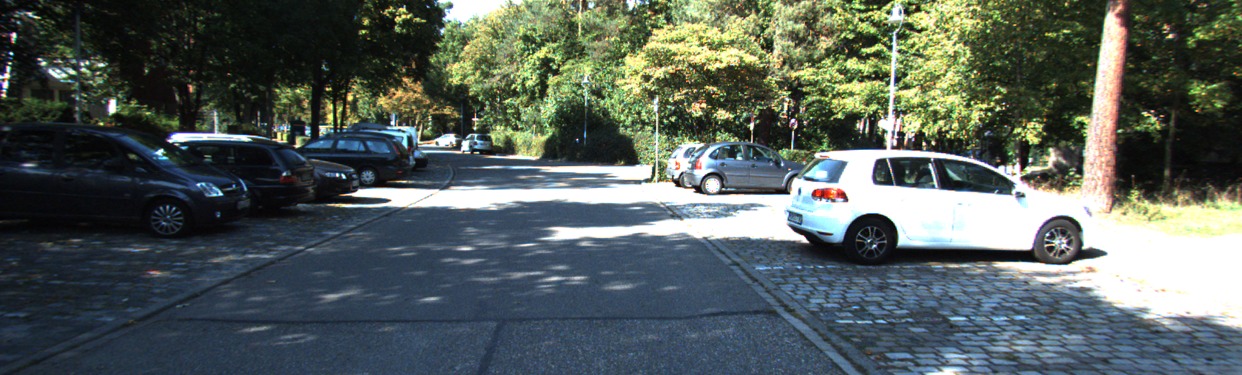}&\includegraphics[width=0.23\linewidth]{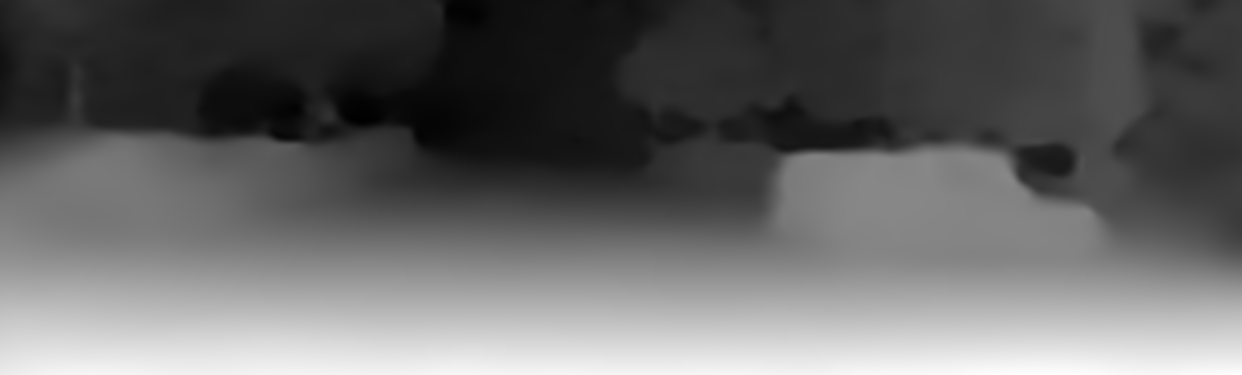}&\includegraphics[width=0.23\linewidth]{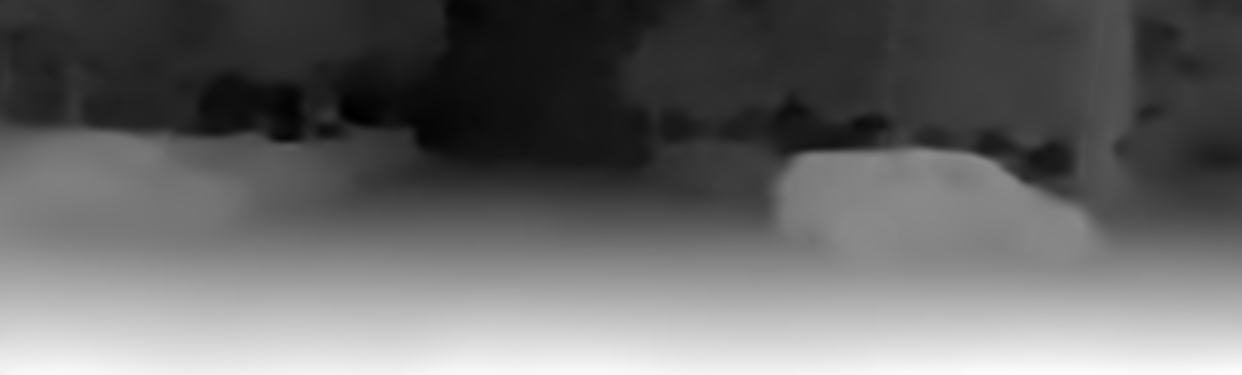}&\includegraphics[width=0.23\linewidth]{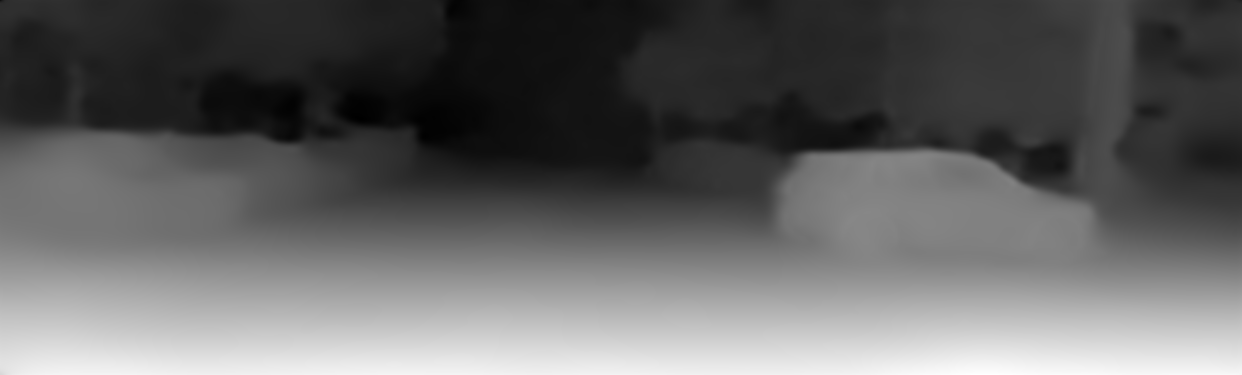}\\
    \vspace{3mm}
    {\hspace{0mm}}
    \includegraphics[width=0.23\linewidth]{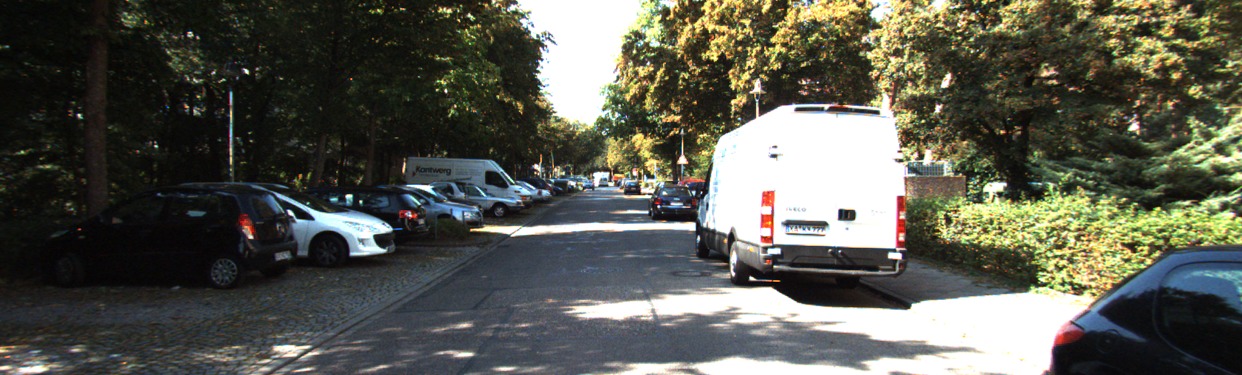}&\includegraphics[width=0.23\linewidth]{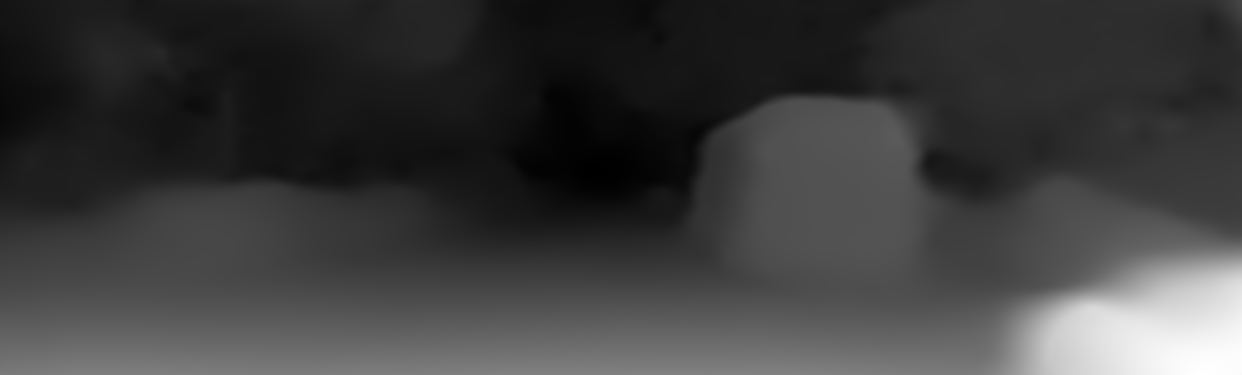}&\includegraphics[width=0.23\linewidth]{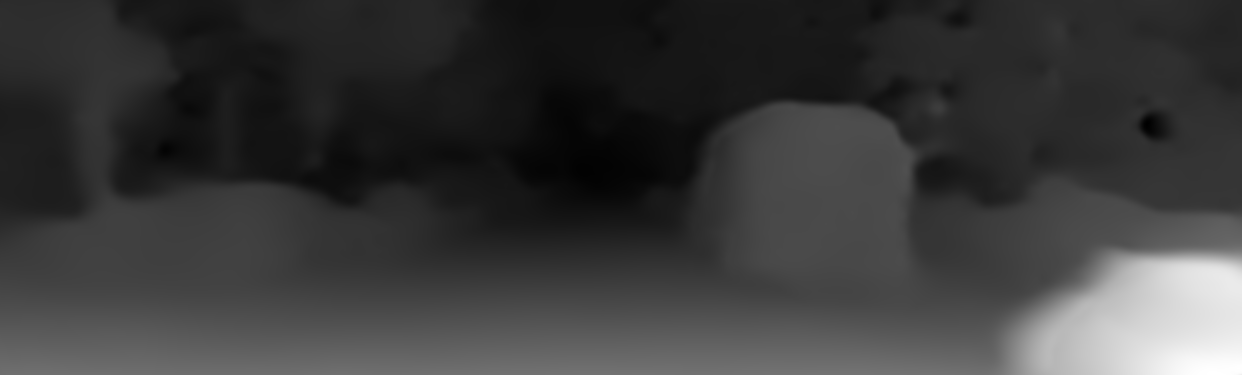}&\includegraphics[width=0.23\linewidth]{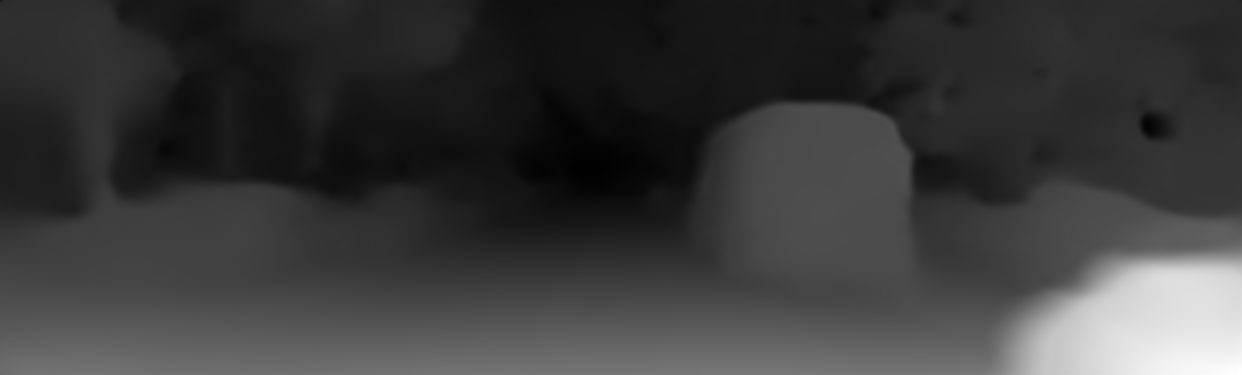}\\
    \vspace{3mm}
    {\hspace{0mm}}
    \includegraphics[width=0.23\linewidth]{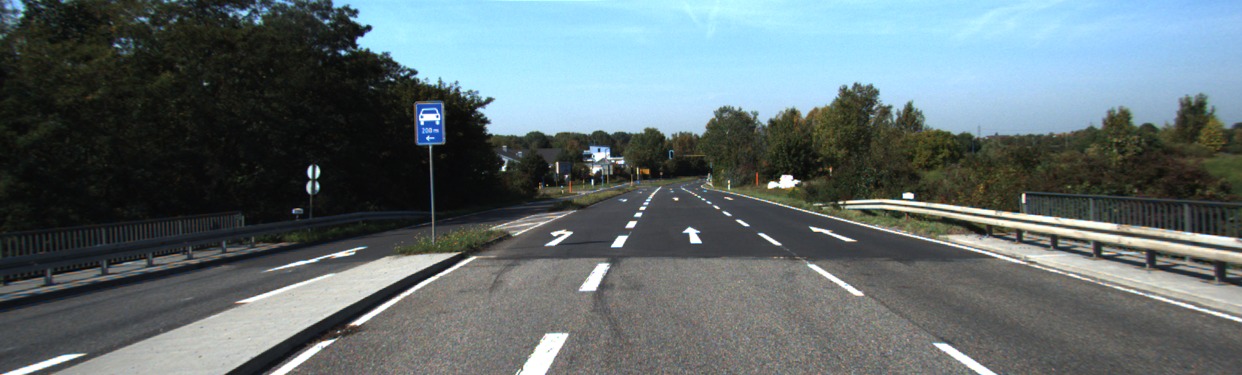}&\includegraphics[width=0.23\linewidth]{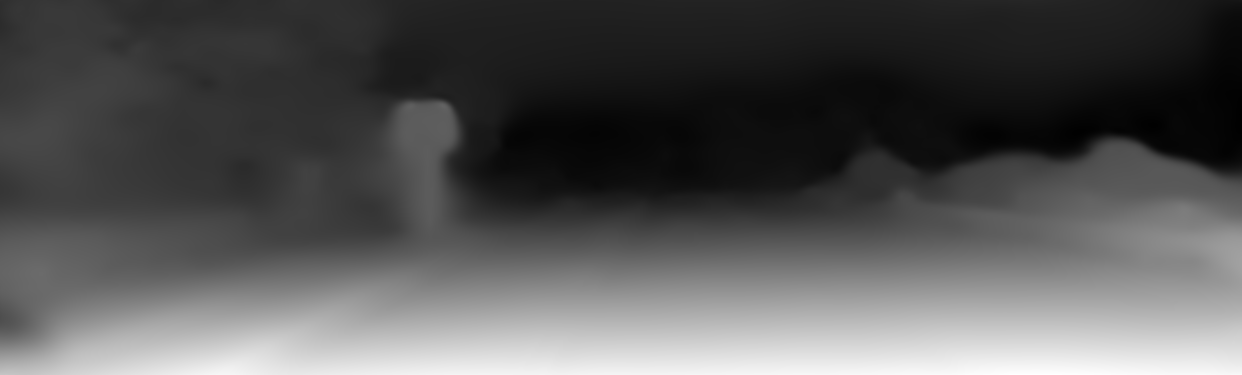}&\includegraphics[width=0.23\linewidth]{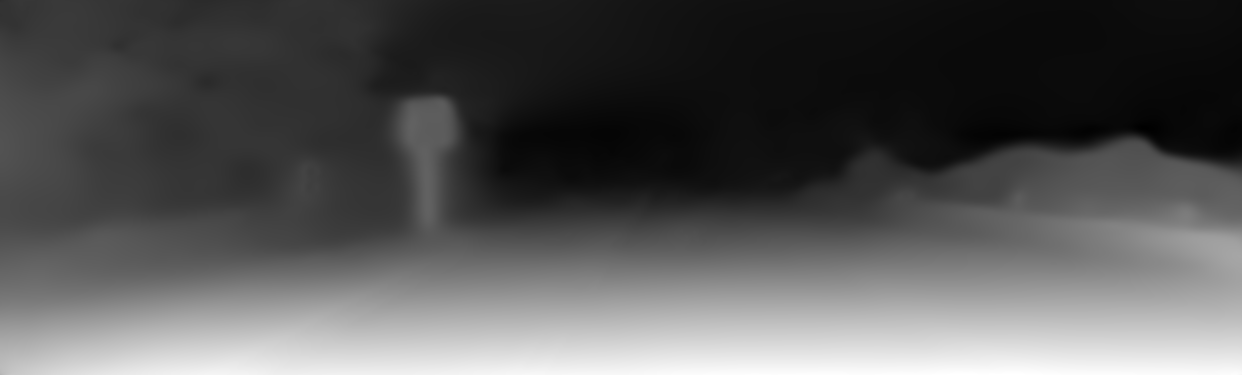}&\includegraphics[width=0.23\linewidth]{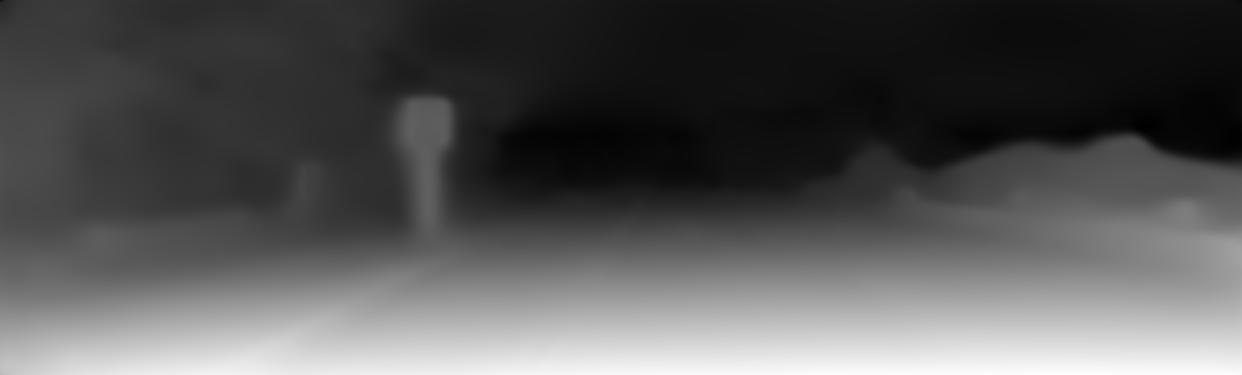}\\
    \vspace{3mm}
    {\hspace{0mm}}
    \includegraphics[width=0.23\linewidth]{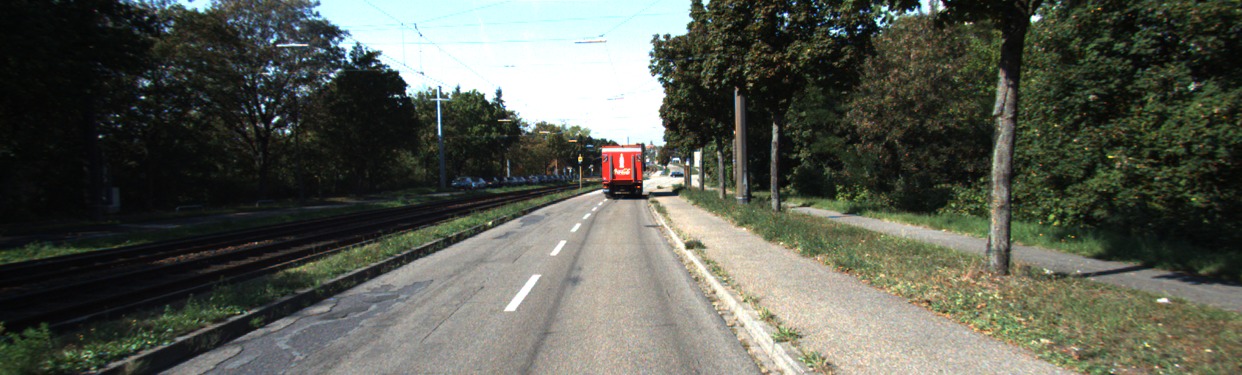}&\includegraphics[width=0.23\linewidth]{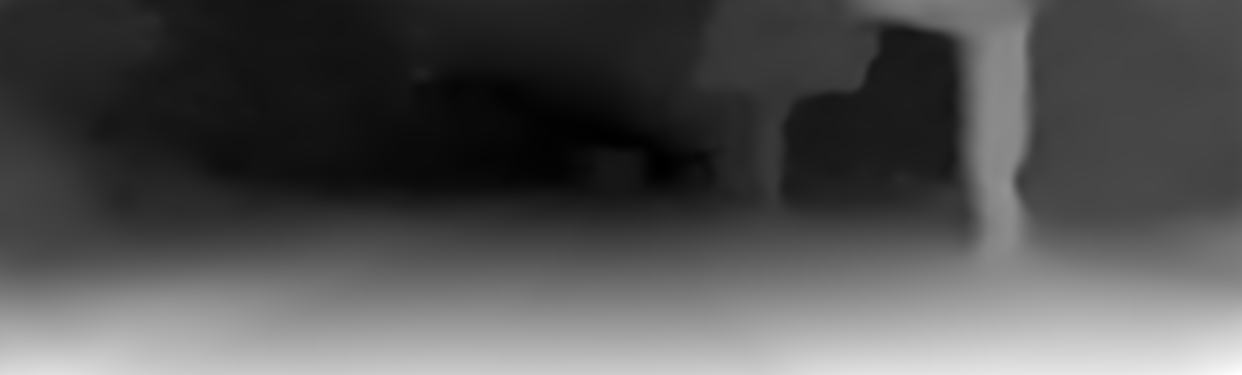}&\includegraphics[width=0.23\linewidth]{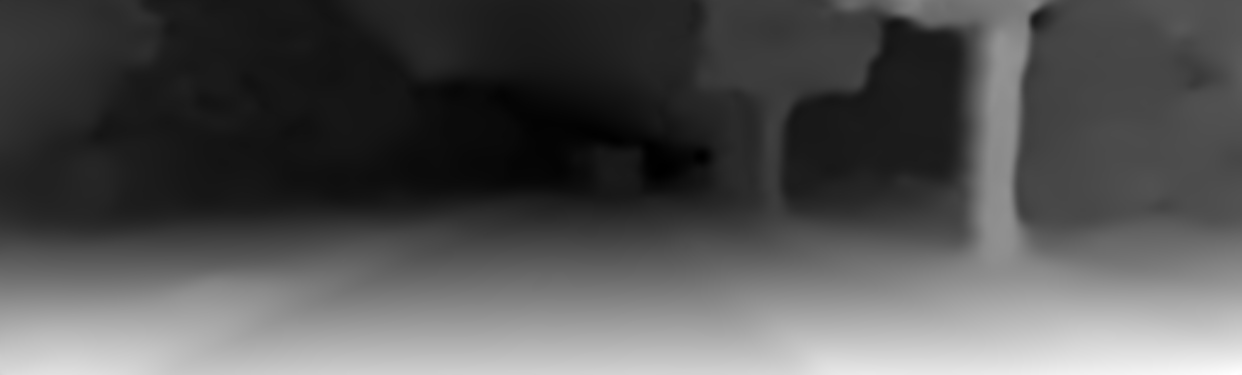}&\includegraphics[width=0.23\linewidth]{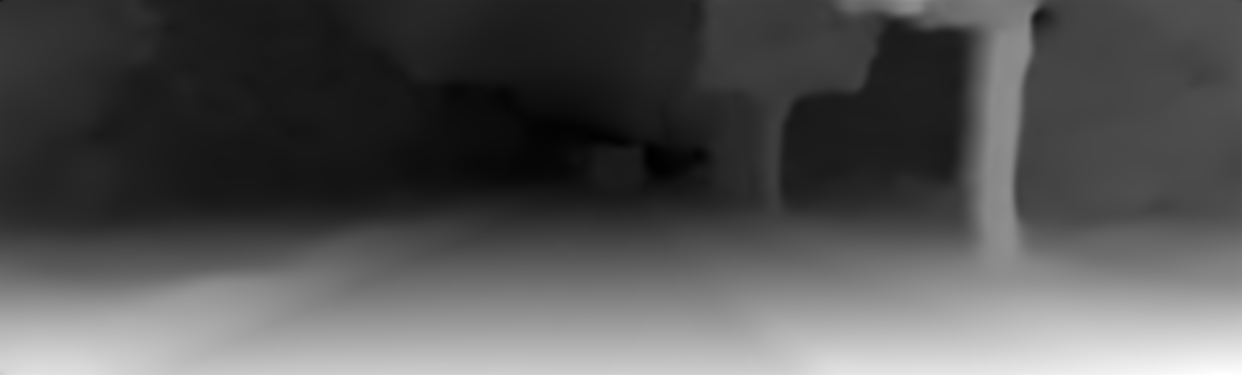}\\
    \vspace{3mm}
    {\hspace{0mm}}
    \includegraphics[width=0.23\linewidth]{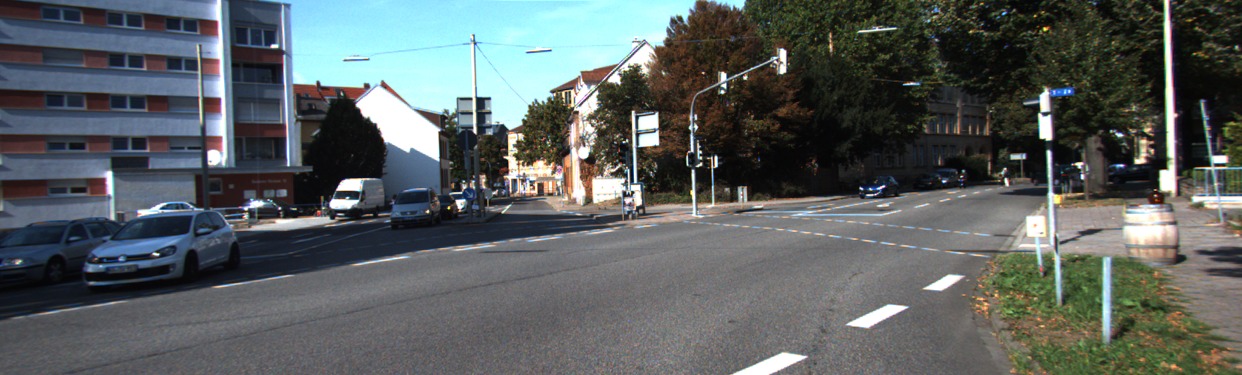}&\includegraphics[width=0.23\linewidth]{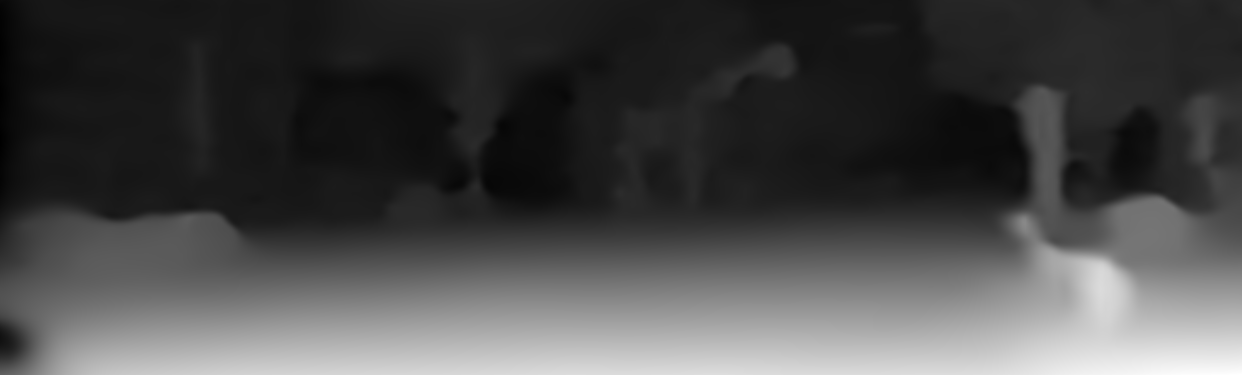}&\includegraphics[width=0.23\linewidth]{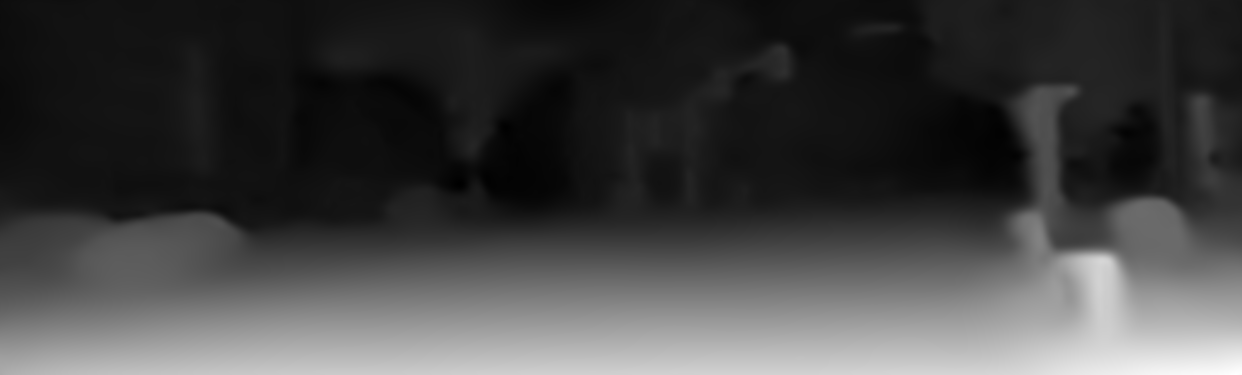}&\includegraphics[width=0.23\linewidth]{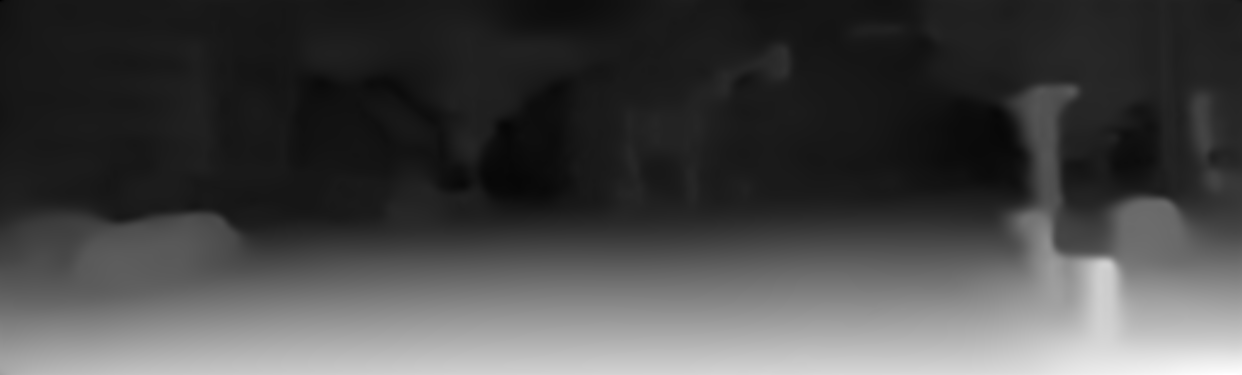}\\
    \vspace{3mm}
    {\hspace{0mm}}
    \includegraphics[width=0.23\linewidth]{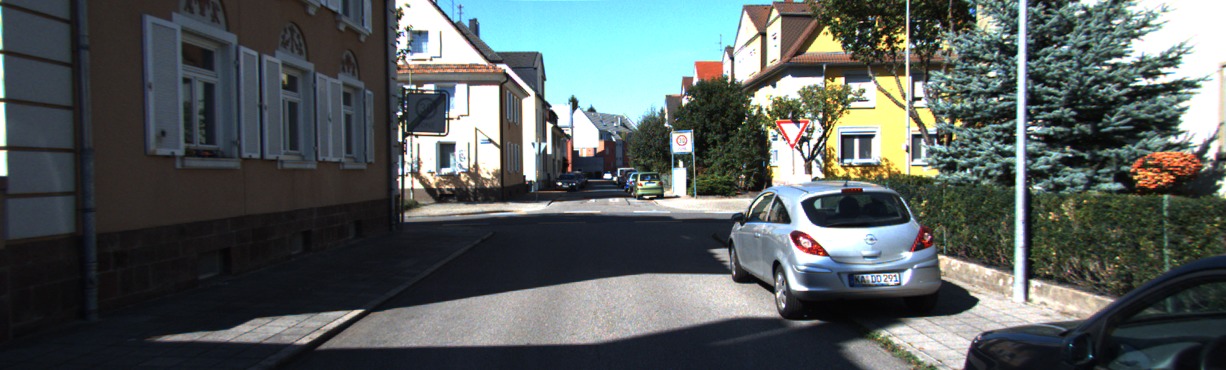}&\includegraphics[width=0.23\linewidth]{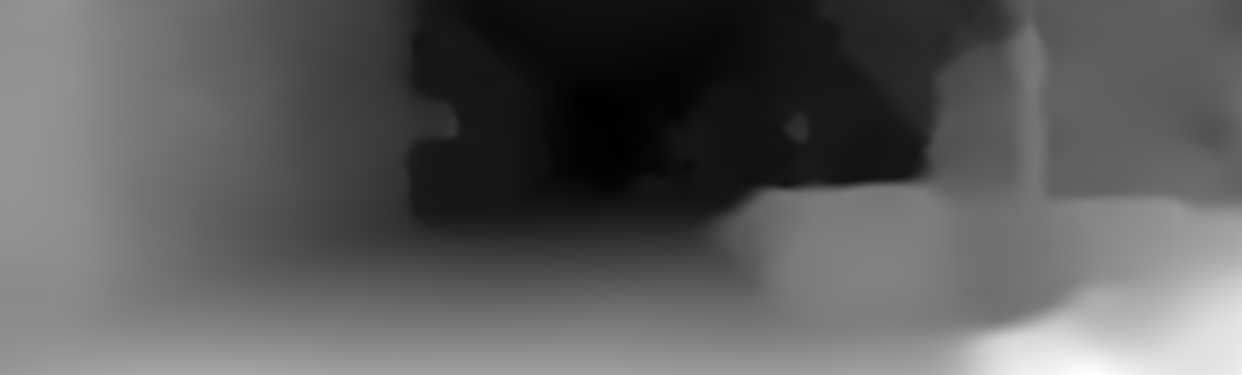}&\includegraphics[width=0.23\linewidth]{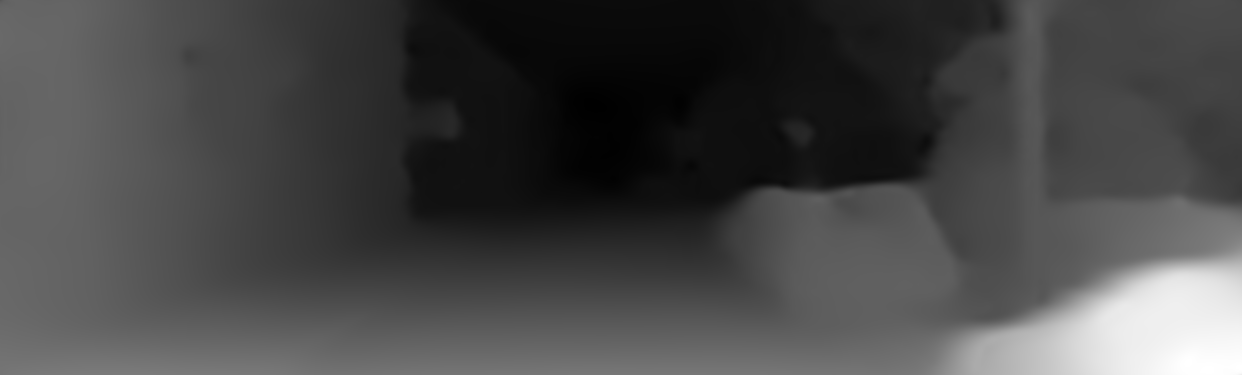}&\includegraphics[width=0.23\linewidth]{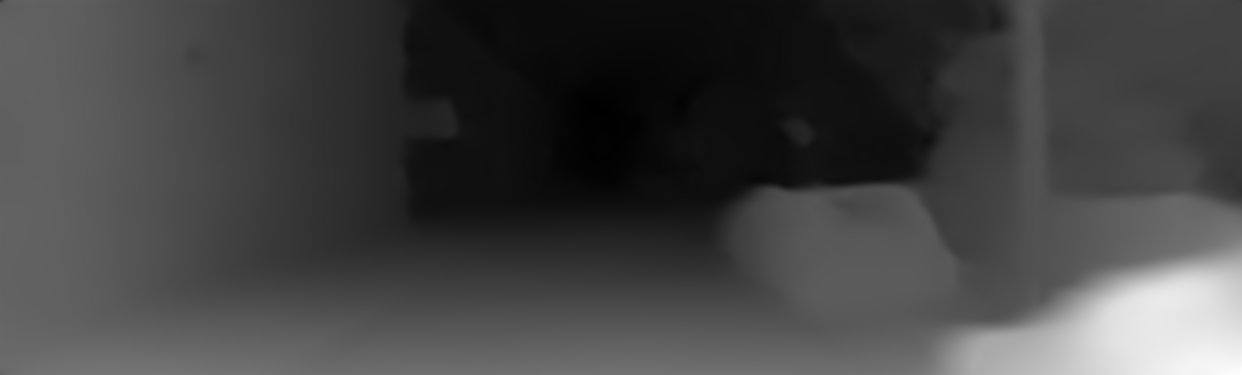}\\
    {\hspace{0mm}}
    \includegraphics[width=0.23\linewidth]{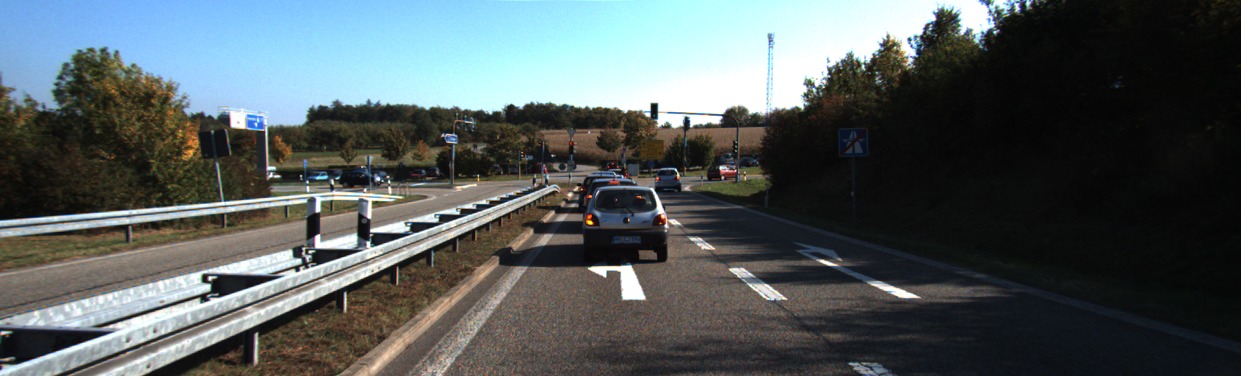}&\includegraphics[width=0.23\linewidth]{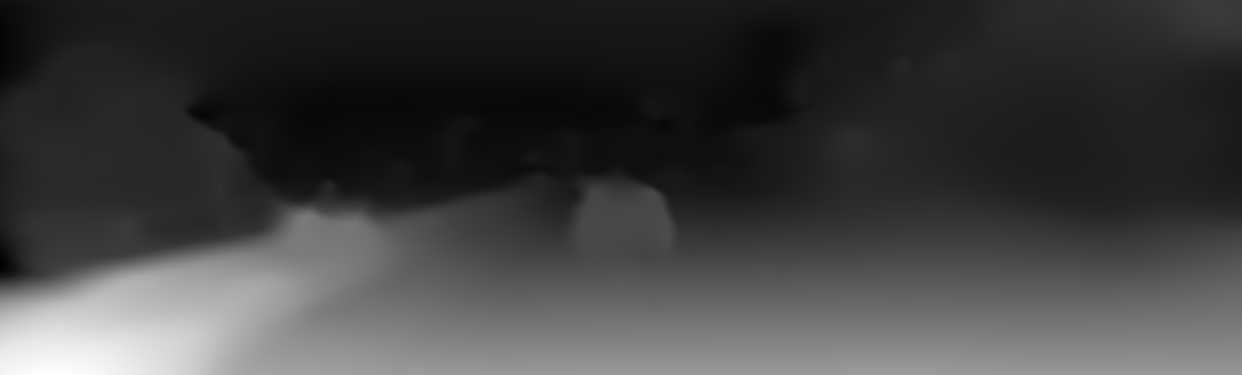}&\includegraphics[width=0.23\linewidth]{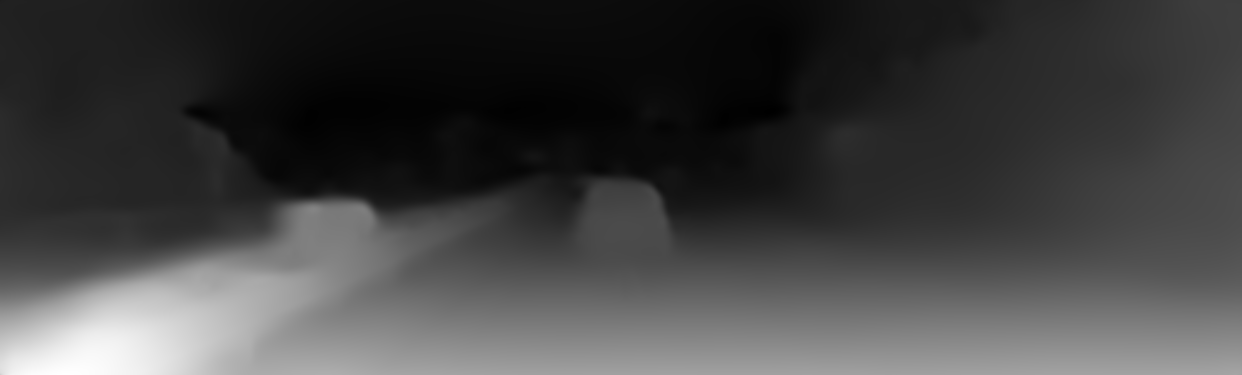}&\includegraphics[width=0.23\linewidth]{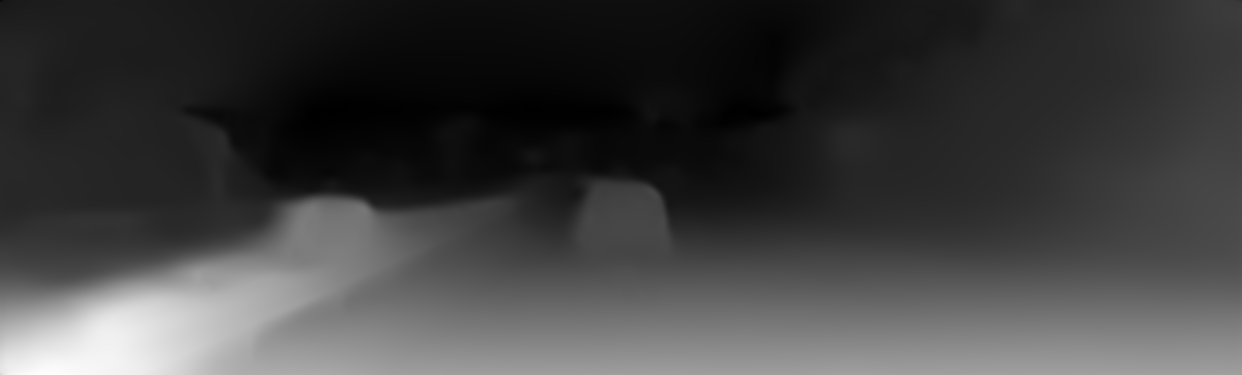}\\
\end{tabular}
\caption{Qualitative results on the Eigen test split. The boundary is more clear and accurate as we add flow pairs and the proposed 2-warp consistency during training.}
\label{fig:comparison}
\end{figure*}

\end{document}